\documentclass[journal,balance]{IEEEtran}

\ifCLASSINFOpdf
\else
\fi

\usepackage{amsmath,amsfonts,amssymb}
\usepackage{color}
\usepackage{tikz}
\usepackage{balance}
\usepackage{verbatim}
\usepackage{hyperref}
\usepackage{url}
\usepackage{placeins}
\usepackage{multirow,hhline}

\newcommand{\ru}    {\rule{0mm}{4mm}}
\newcommand{\ls}    {\hspace{2mm}}
\newcommand{\minus} {\!-\!}
\newcommand{\plus}  {\!+\!}

\newcommand{\be}    {\begin{equation}}
\newcommand{\ee}    {\end{equation}}

\newcommand{\z}     {{\bf z}}
\newcommand{\x}     {{\bf x}}

\newcommand{\gammaf} [1]{\Gamma   \left(#1 \right)}

\newcommand{\psif}   [2]{\psi   ^{\left(#1 \right)} \left(#2 \right)}

\newcommand{\local}{E-Lee}
\newcommand{\bmtd} {BM3D}
\newcommand{\fans} {FANS}
\newcommand{\ppb}  {PPB}
\newcommand{\gbf}  {GBF}
\newcommand{\gnlmA}{GNLM1}
\newcommand{\gnlmC}{GNLM2}

\begin{document}

\title{Guided patch-wise nonlocal {SAR} despeckling}

\author{
Sergio~Vitale, Davide~Cozzolino, Giuseppe~Scarpa, Luisa~Verdoliva, and~Giovanni~Poggi
\thanks{S.~Vitale is with the Engineering Department of University Parthenope, Naples, Italy. e-mail: sergio.vitale@uniparthenope.it.
The other authors are with the Department of Electrical Engineering and Information Technology,
University Federico II, Naples, Italy. e-mail: \{firstname.lastname\}@unina.it.}
}

\maketitle

\begin{abstract}
We propose a new method for SAR image despeckling which leverages information drawn from co-registered optical imagery.
Filtering is performed by plain patch-wise nonlocal means, operating exclusively on SAR data.
However, the filtering weights are computed by taking into account also the optical guide, which is much cleaner than the SAR data, and hence more discriminative.
To avoid injecting optical-domain information into the filtered image,
a SAR-domain statistical test is preliminarily performed to reject right away any risky predictor.
Experiments on two SAR-optical datasets prove the proposed method to suppress very effectively the speckle, preserving structural details, and without introducing visible filtering artifacts.
Overall, the proposed method compares favourably with all state-of-the-art despeckling filters, and also with our own previous optical-guided filter.
\end{abstract}


\IEEEpeerreviewmaketitle

\section{Introduction}

Remote sensing imagery represents nowadays an invaluable source of information for the analysis of the Earth's state.
Among the many types of sensors, synthetic aperture radar (SAR) systems are especially precious,
since they observe features that are not captured by optical sensors, and acquire data on the target scene independently of the weather and lighting conditions.
SAR images are routinely exploited in many key applicative fields,
like, for example, the analysis of the environment \cite{Sarker2013, Irwin2018} or urban planning \cite{He2006, Gamba2011}.
Unfortunately, they are severely impaired by the presence of speckle, caused by the coherent nature of the scattering phenomena.
Speckle noise strongly degrades the quality of SAR images,
thereby affecting the performance of subsequent automated tasks, like segmentation \cite{Deng2005, Wang2014} or classification \cite{Dekker2003, Popescu2016},
and causing problems even to human interpreters.

To tackle this problem,
the scientific community has produced a major research effort on SAR despeckling in the last decades \cite{Argenti2013}.
A large number of methods have been proposed,
working in the spatial domain \cite{Lee1983, Lopes1990, Lee2009},
using wavelet transform \cite{Guo1994, Achim2003, Argenti2008},
sparse representations \cite{Foucher2008, Ozcan2016, Tabti2018},
variational approaches \cite{Denis2009, Feng2014, Zhao2015} and, very recently,
deep learning \cite{Chierchia2017, Wang2017}.
As of today, however, the most successful approach to SAR despeckling appears to be the nonlocal paradigm \cite{Deledalle2014, DiMartino2014},
which has produced powerful and widespread methods such as PPB \cite{Deledalle2009} and SAR-BM3D \cite{Parrilli2012}.
Key to this success is the ability to recognize ``similar'' pixels, that is, pixels characterized by the same underlying signal.
This allows, for each target pixel, to single out its best predictors in the whole image, and use them to perform reliable estimation.
Therefore, in nonlocal filtering the main issue is how to find such good predictors.
This problem is usually addressed by using suitable patch-based similarity measures \cite{Deledalle2012}, leveraging the contextual information conveyed by patch-wise analysis.
However, speckle impacts also this process, reducing the ability to find good predictors and, eventually, impairing the filtering performance.

\begin{figure}
	\centering\footnotesize
	\setlength{\tabcolsep}{1pt}
	\begin{tabular}{c@{\hspace{1mm}}c}
		\includegraphics[width=4.0cm]{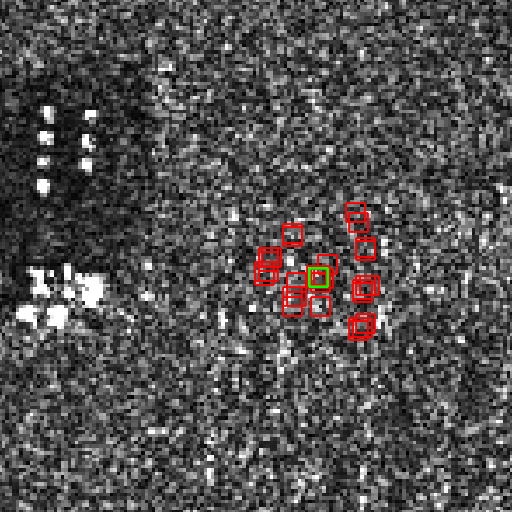} &
		\includegraphics[width=4.0cm]{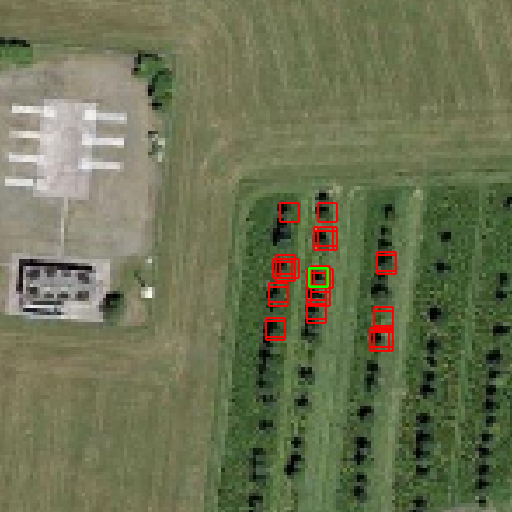} \\
		\includegraphics[width=4.0cm]{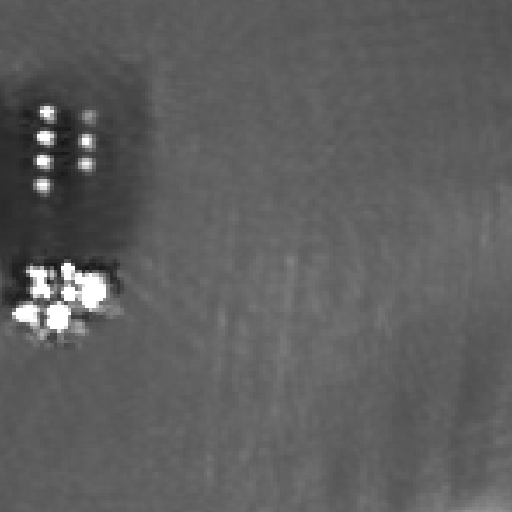} &
		\includegraphics[width=4.0cm]{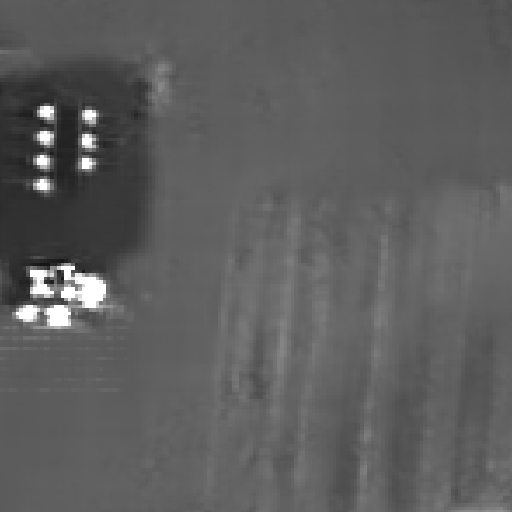}
	\end{tabular}
    \caption{Optical-guided nonlocal filtering.
    Top: co-registered SAR and optical images with a target patch (green) and its best matches (red).
    The similarity measure singles out the best predictors in the optical image, not so in the SAR image.
    Bottom: Output of a conventional SAR-domain nonlocal filter (PPB), and of the proposed optical-guided nonlocal filter.
    The optical guide allows to better preserve all image structures, without introducing filtering artifacts.}
    \label{fig:best_predictors}
\end{figure}

We highlight the limits of the nonlocal approach for SAR despeckling with the help of Fig.\ref{fig:best_predictors}
which shows a single-look SAR image (top-left) together with an optical image of the same scene co-registered with it (top-right).
The SAR image carries precious information on the scene which is not available in the optical bands.
Nevertheless the signal of interest is overwhelmed by speckle noise:
the scene structure is hardly visible, and the boundaries between different land covers can be barely detected.
This impacts heavily on nonlocal filtering, preventing the correct localization of the best predictors.
As an example, for the target patch marked by a green box in the figure, the selected predictors (red boxes) are dominated by speckle and hence spread all around the target.
With such a poor selection of predictors, plain nonlocal filtering can only provide unsatisfactory results (bottom-left).
On the optical image, however, which is virtually noiseless, finding good predictors is very easy.
Now the selected patches (red boxes) exhibit clearly a signal content similar to the target, and are very likely the best possible predictors.
Based on these observations, we decided to leverage optical data to improve nonlocal SAR despeckling,
obtaining promising results, as shown in the figure (bottom-right).

Of course, data fusion is nothing new in remote sensing.
The large abundance of imagery from sensors of different types offers a wealth of opportunities \cite{Schmitt2016, Joshi2016, Schmitt2017, Bahmanyar2018}
that can be exploited for many remote sensing applications \cite{Corbane2008, Laurin2013, Errico2015, Idol2015}.

\begin{figure}
	\centering\footnotesize
	\setlength{\tabcolsep}{1pt}
	\begin{tabular}{c@{\hspace{1mm}}c}
		\includegraphics[width=4.0cm]{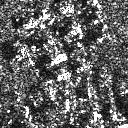} &
		\includegraphics[width=4.0cm]{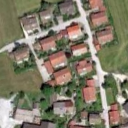}
	\end{tabular}
    \caption{Differences between co-registered SAR and optical images.
    In the presence of man-made structures the images are profoundly different, and using optical data to guide despeckling may cause serious filtering artifacts.}
    \label{fig:structural_differences}
\end{figure}

However, using optical data to support SAR despeckling requires great care.
The example of Fig.\ref{fig:structural_differences} provides some insight into this point.
In fact, despite the careful co-registration, and the obvious correspondence of the observed scene,
important differences exist between optical and SAR images.
These regard not only the signal amplitudes, which show no obvious relationship due to the completely different imaging mechanisms,
but also the signal structure, especially in the presence of man-made objects and regions characterized by a significant orography (not present in this example).
Therefore, while optical data can be certainly helpful to guide the despeckling process,
there is the risk to inject alien information into the filtered SAR image, generating annoying artifacts.

Based on these concepts,
in \cite{Verdoliva2015} we proposed a nonlocal despeckling technique for SAR images, driven by co-registered optical images.
Within the frame of a generalized bilateral filtering, optical data were used to properly weight predictor pixels for the current target.
To prevent the injection of alien optical structures,
the SAR image was preliminary classified, and the optical guide was used only in low-activity areas, switching to a full SAR-domain technique in high-activity areas.

In this work (a preliminary version of which was presented in \cite{Gaetano2017}) we keep pursuing the same general approach but propose a much more effective and simple optical-guided SAR despeckling method.
We replace the pixel-wise bilateral filter of \cite{Verdoliva2015} with patch-wise nonlocal means.
Moreover, to avoid optical-related artifacts, we use a simple statistical test which discards unreliable predictors on the fly, during the filtering process.
Extensive experiments on real-world imagery prove the potential of the proposed method,
also in comparison with state-of-the-art reference methods and with our own previous proposal.
In addition, by avoiding the preliminary classification phase and the external complementary filter, the method is much faster and easy to use than \cite{Verdoliva2015}.
Note that, with respect to our conference paper \cite{Gaetano2017},
we introduce here a reliability test which allows us to despeckle effectively also high-activity areas, keeping all available information and removing only bad predictors.
Moreover, we perform a theoretical analysis of the proposed test, and carry out a much deeper experimental analysis of performance.

In the rest of the paper
after recalling previous work (Section 2),
we describe the proposed solution (Section 3),
study the effect of key parameters on performance (Section 4),
discuss experimental results (Section 5),
and eventually draw conclusions (Section 6).
The SAR-domain distance used in the reliability test is analyzed in Appendix A.

\section{Previous work}

The optical-guided despeckling paradigm was first proposed in \cite{Verdoliva2015}.
It was observed, there, that a virtually noiseless optical image, 
co-registered with the target SAR image, can provide precious information to support the speckle removal process.
Although SAR and optical images refer to completely different imaging mechanisms,
and hence there is no relationship between their signal amplitudes, they share important structural information.
A boundary between two fields, for example, keeps the same geometry in both the optical and the SAR image.
This structural information can be exploited by means of guided filtering.
However, one must also be aware that such a structural similarity does not always hold.
This is the case of man-made areas, for example,
characterized by intense double reflection lines in SAR images that have no correspondence in optical images.
Therefore, care must be taken not to generate filtering artifacts due to the optical guide.

\begin{figure}[t]
\centering
    \includegraphics[width=0.80\columnwidth]{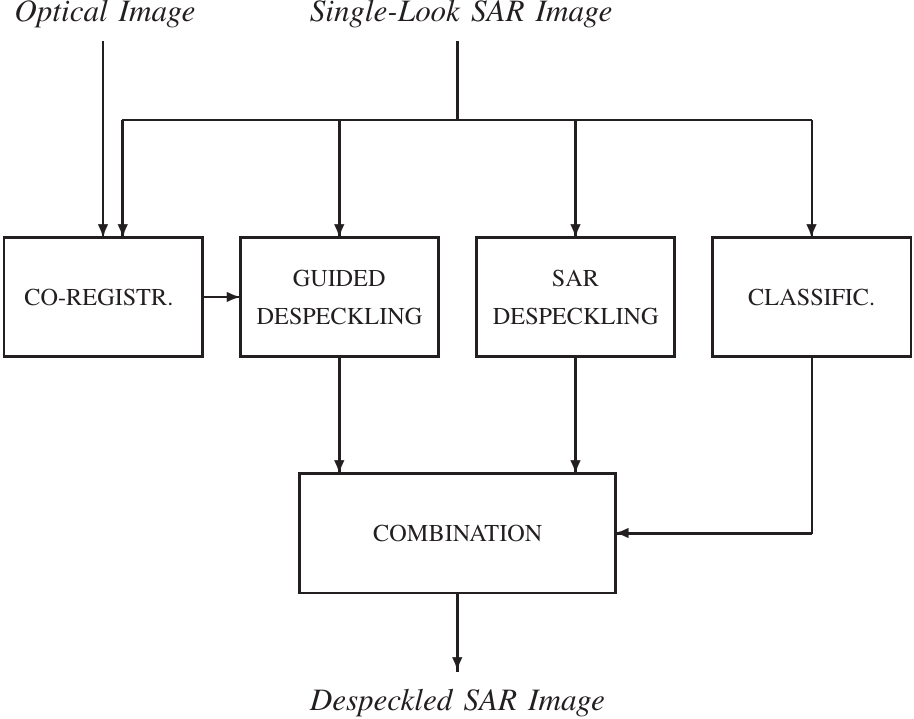}
    \caption{Block diagram of the optical-guided despeckling method of \cite{Verdoliva2015}.}
\label{fig:flowchart}
\end{figure}

In \cite{Verdoliva2015} the problem was solved by introducing a preliminary soft classification phase.
A high-level scheme of the method is shown in Fig.\ref{fig:flowchart}.
The single-look input SAR image is filtered twice:
by means of a guided despeckling tool, leveraging the co-regirested optical image, and
by means of a conventional SAR-domain despeckling filter.
A soft classifier \cite{Gragnaniello2016} distinguishes between low-activity and high-activity regions, the latter possibly related to man-made areas where optical and SAR geometries differ.
The output image is then obtained as a linear combination of the two filtered images, with weights given by the continuous-valued classifier.
In low-activity areas, the guided filter prevails, while the opposite happens in high-activity areas.

For the despeckling of high-activity areas we used SAR-BM3D \cite{Parrilli2012, Cozzolino2014},
which is known \cite{DiMartino2014} to guarantee a good preservation of fine details and man-made structures.
The guided filter for low-activity areas, instead, is a generalization of the bilateral filter \cite{Tomasi1998}.
Assuming the usual multiplicative model, $z(t)=x(t)u(t)$,
with $x(t)$ and $z(t)$ the true and observed intensity values at pixel $t$, and $u(t)$ a Gamma random variable (RV) modeling the speckle,
the estimated intensity $\widehat{x}(t)$ is given by the weighted average of predictors $z(s)$ drawn from a small neighborhood $\Omega(t)$ of $t$:
\begin{equation}
    \widehat{x}(t) = \sum_{s \in \Omega(t)} w(s,t) \, z(s)
    \label{eq:pixelwise_NLM}
\end{equation}
The weight associated with the predictor at location $s$ is computed as
\begin{eqnarray}
    w(s,t) & = & C \exp \left\{ -\alpha \| s \minus t \|^2 +\right. \\
           &   &        \left.  -\lambda_O d_O[o(s),o(t)] -\lambda_S d_S[z(s),z(t)] \right\} \rule{0mm}{4mm} \nonumber
    \label{eq:pixelwise_weights}
\end{eqnarray}
where $o(\cdot)$ indicates the vector-valued optical data, and $C$ is a normalizing constant.
The weights depend both on spatial distance, $\| s \minus t \|^2$, like in a ordinary Gaussian filter, and amplitude distances in the SAR domain, $d_S[z(s),z(t)]$, and in the optical domain, $d_O[o(s),o(t)]$.
A simple Euclidean norm is used for the optical-domain distance,
while in the SAR-domain we use the dissimilarity measure (loosely referred to as distance in the following) proposed in \cite{Deledalle2009} for multiplicative noise.
The weights $\alpha, \lambda_O$ and $\lambda_S$ are set by preliminary experiments on training data.
When $\lambda_O=0$ a simple bilateral filter in the SAR domain is obtained.
On the contrary, if $\alpha=\lambda_S=0$, the weights depend only on the optical-domain distances.
However, it is worth underlining that in the filtering procedure there is no leakage of optical data in the filtered output.
The filtered value in Eq.(\ref{eq:pixelwise_NLM}) is a linear combination of exclusively SAR-domain original values, and optical data impact only on their weighting.
Likewise, the soft classifiers uses only SAR data as input.
Hence, the optical guide only helps
locating predictors that are most similar to the target or, under a different perspective,
de-emphasizing the contributes of predictors that are not really similar to the target despite their low spatial and SAR-domain distances.

\section{Proposed method}

This work introduces two major improvements with respect to \cite{Verdoliva2015}, consisting in
\begin{itemize}
\item   replacing the pixel-wise generalized bilateral filter with patch-wise nonlocal means;
\item   using a reliability test to reject poor predictors on the fly and prevent structural leakages from optical data.
\end{itemize}
The resulting filter, besides providing a much better performance, is much simpler and easy to use,
since we remove altogether the activity-based classifier, and do not need external filters to manage high-activity areas.

\subsection{Going patch-wise}
\label{sec:going}

In recent years, there has been a steady trend towards patch-based processing for SAR imagery \cite{Deledalle2014}.
Patch-wise nonlocal means, in particular, is well known to significantly outperform the pixel-wise version.
The key idea is to compute a large number of estimates of the same pixel, which are eventually aggregated to improve accuracy.
This is obtained by applying the nonlocal weighted average to all pixels of a patch, not just its center.

Let us consider a target patch, $\z(t)=\{z(t+k), k \in \mathbb{P}$\},
where $t$ is an anchor pixel (for example, the patch center), and $\mathbb{P}$ indicates the set of spatial offsets with respect to $t$.
Then, an estimate of the clean patch $\x(t)$ is obtained through a patch-wise nonlocal average
\begin{equation}
    \widehat{\x}(t) = \sum_{s\in\Omega(t)} w(s,t) \z(s)
    \label{eq:patchwise_NLM}
\end{equation}
where the weights $w(s,t)$ depend on a suitable patch-wise similarity measure.
This is the same expression as in Eq.(\ref{eq:pixelwise_NLM}), except that it now involves all pixels in the target patch, namely
\begin{equation}
    \widehat{x}(t + k) = \sum_{s\in\Omega(t)} w(s,t) \, z(s + k) \;\;\; \forall k \in \mathbb{P}
    \label{eq:patchwise_NLM_expanded}
\end{equation}
Since a pixel belongs to multiple target patches, it will be estimated several times, allowing for the eventual aggregation of all estimates.

The weights are now computed as
\begin{equation}
    w(s,t) = C\exp \left\{-\lambda[\gamma d_S(s,t)+(1-\gamma)d_O(s,t)] \right\}
\end{equation}
where $d_S(s,t)$ and $d_O(s,t)$ are suitable SAR-domain and optical-domain distances,
$\gamma \in [0,1]$ is a parameter that balances their contribution,
and $\lambda$ is a parameter which determines how fast weights decay as a function of the distance, and hence impacts on the sharpness/smoothness of the filtered image.
Note that, unlike in \cite{Verdoliva2015}, the weights depend only on signal amplitudes, both SAR and optical, not on spatial distances.

For the SAR domain, we use a normalized and slightly modified version of the distance proposed in \cite{Deledalle2009}
\begin{equation}
    d_S(s,t) = \frac{1}{\mu_D}\frac{1}{N} \sum_{k \in \mathbb{P}} \log \left[ \frac{z(s+k)+z(t+k)}{2 \sqrt{z(s+k)z(t+k)}} \right]
\label{eq:dSAR}
\end{equation}
where $N=|\mathbb{P}|$ is the patch size, and $\mu_D$ (described in more detail in Subsection III.C and in the Appendix) is the mean of the single-pixel distance under same-signal hypothesis.
The normalization ensures that, in strictly homogeneous regions,
predictor patches have unitary average distance from the target, say $\mu_P=1$, with standard deviation $\sigma_P=\sigma_D/\mu_D\sqrt{N}$.
For the optical domain, instead, we consider the normalized Euclidean distance
\begin{equation}
    d_O(s,t) = \frac{1}{MN} \sum_{i=1}^M \sum_{k \in \mathbb{P}} [o_i(s+k) - o_i(t+k)]^2
\label{eq:doptical}
\end{equation}
with $M$ the number of bands.


\begin{figure}
	\centering\footnotesize
	\setlength{\tabcolsep}{1pt}
	\begin{tabular}{c@{\hspace{1mm}}c}
		\includegraphics[width=4.4cm]{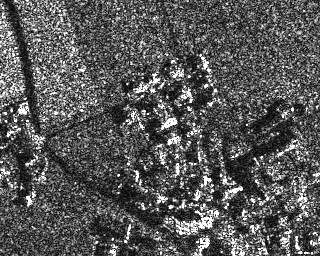} &
		\includegraphics[width=4.4cm]{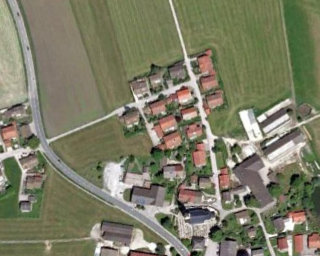} \\
		\includegraphics[width=4.4cm]{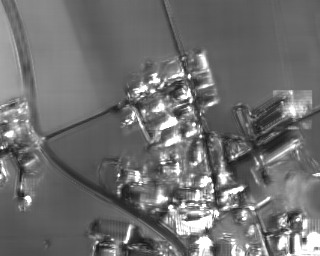} &
		\includegraphics[width=4.4cm]{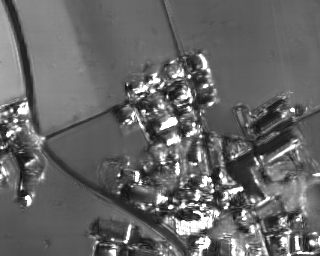} \\
		\includegraphics[width=4.4cm]{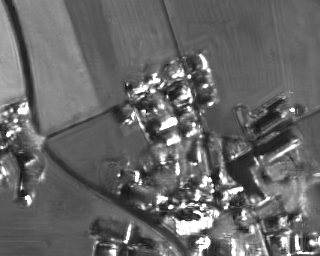} &
		\includegraphics[width=4.4cm]{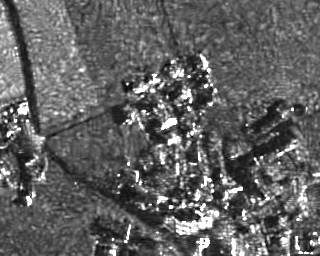}
	\end{tabular}
    \caption{Top: co-registered SAR and optical images.
    Center: strong reflectors cause filtering artifacts in their whole neighborhood (left);
    these are removed thanks to the test on the SAR distance (right).
    Bottom: limiting the number of predictors improves resolution in low-activity areas, with a final result (left)
    which compares quite favourably with conventional SAR-domain filters, such as SAR-BM3D (right).}
    \label{fig:NLM_ghosts}
\end{figure}

\subsection{Discarding unreliable predictors}
\label{sec:discard}

Patch-level processing allows for a much stronger noise suppression than pixel-level processing \cite{Deledalle2014}.
Nonetheless, for our application, patch-level operations entail also some risks.
Consider, for example, a low-activity target patch with some high-activity patches in its neighborhood, $\Omega(t)$, maybe patches with double reflection lines.
Since {\em all} patches in $\Omega(t)$ are averaged to estimate the clean target,
the structures observed in the predictors will be reproduced in the estimate, with an attenuation that depends on optical-domain and SAR-domain distances.
Hence, insufficient attenuation of high-activity patches will generate visible artifacts.
This can easily happen when
the SAR-domain distance is not too large ({\it e.g.} just a few double-reflection pixels in the predictor) and the optical-domain distance is not particularly discriminative.
Fig.\ref{fig:NLM_ghosts} shows a clear example of this behavior.
A few high-intensity pixels, included in the estimation of surrounding patches, produce disturbing artifacts in a large area of the filtered image.

Of course, to reduce such artifacts, one can modify the filter's parameters, increasing the relative weight of SAR-domain vs. optical-domain distance.
However, this would reduce the benefit of the optical guide in other areas where such artifacts do not occur.
To cope with this problem, in \cite{Verdoliva2015} we decided to avoid patch-level processing altogether and to treat differently low-activity and high-activity areas.
Here, we use a more effective solution.
We keep using patch-wise processing but, for each target patch, carry out a preliminary test to single out unreliable predictors and exclude them altogether from the nonlocal average.
That is, we perform the nonlocal average of Eq.(\ref{eq:patchwise_NLM}) replacing $\Omega(t)$ with a new set $\Omega'(t)$ such that
\begin{equation}
    \Omega'(t) = \{s \in \Omega(t): d_S(s,t)<T\}
\end{equation}
With this solution, we are free to select the filter parameters that optimize performance in low-activity areas.
Moreover, by removing problematic patches beforehand, we can keep using patch-wise averages also in high-activity areas, with clear benefits in terms of speckle suppression.
Only in the extreme case in which no predictor is reliable, maybe due to the presence of corner reflectors, the target patch is not filtered at all,
which makes sense in this condition.
However, since each pixel belongs to many patches, it is still likely that many individual pixels will be filtered anyway.

Of course, the success of this strategy depends on the discriminative power of the SAR-domain distance, and on a suitable selection of the threshold.
These aspects are analyzed in the following subsection.
In the example of Fig.\ref{fig:NLM_ghosts}, however, it is already clear that this simple test impacts heavily on the quality of filtered image.

A further problem, besides filtering artifacts, is image oversmoothing, and the consequent loss of resolution.
This effect is especially visible at the boundary between fields, where signal differences are small both in the optical and SAR domains.
In this situation,
the weights depend only mildly on the signal, and more strongly on the intense SAR-domain speckle, causing an incoherent averaging of patches and ultimately the loss of details.
In Fig.\ref{fig:NLM_ghosts}, for example, a small road between two fields goes completely lost.
These losses can be reduced by limiting the maximum number of predictors to $S_0<S$ patches, with $S$ the search area size, choosing those with smallest optical-domain distance.
Therefore, the new set $\Omega''(t) \subseteq \Omega'(t)$, has cardinality
\begin{equation}
    |\Omega''(t)| = \min\{|\Omega'(t)|,S_0\}
\end{equation}
Thanks to this limitation, in homogeneous areas of the image many irrelevant patches are excluded from the average, emphasizing fine details that would be lost otherwise.
Instead, in high-activity areas this limitation has no effect, since most predictors are already discarded by the SAR-domain test.
Back to Fig.\ref{fig:NLM_ghosts},
we see that this further limitation allows recovering the road as well as many other details, with no big loss in terms of speckle rejection.
The final result of filtering shows both strong speckle suppression and good detail preservation,
comparing favourably with state-of-the-art filters.

\subsection{SAR-domain distance and reliability test}

Our filtering strategy founds heavily on the reliability test's capacity of rejecting bad predictors.
Therefore, it is worth investigating in more depth the SAR-domain distance, also to gain sensitivity on how to set the decision threshold $T$.

Let us focus, for the time being, on the {\it pixel-wise} SAR-domain distance
\begin{equation}
    D[z(s),z(t)] = \log \left[ \frac{z(s)+z(t)}{2\sqrt{z(s)\,z(t)}} \right]
\end{equation}
This is not exactly the distance proposed in \cite{Deledalle2009} because of the extra 2 at the denominator,
which we include to ensure zero distance for $z(s)=z(t)$.

Based on measured distances, we would like to identify pixels that have a similar signal component as the target, that is, $x(s) \simeq x(t)$.
Of course, since the observed intensity, $z(s)=x(s)u(s)$, depends also on the speckle, the distance depends on the speckle too.
Let us consider two limiting cases, {\it (i)} speckle-free data, and {\it (ii)} homogeneous signal.

\begin{figure}[t]
    \centering
    \begin{tabular}{cc}
    \setlength{\tabcolsep}{1pt}
        \includegraphics[trim={2.4cm 0 2.4cm 0},clip,width=4.0cm]{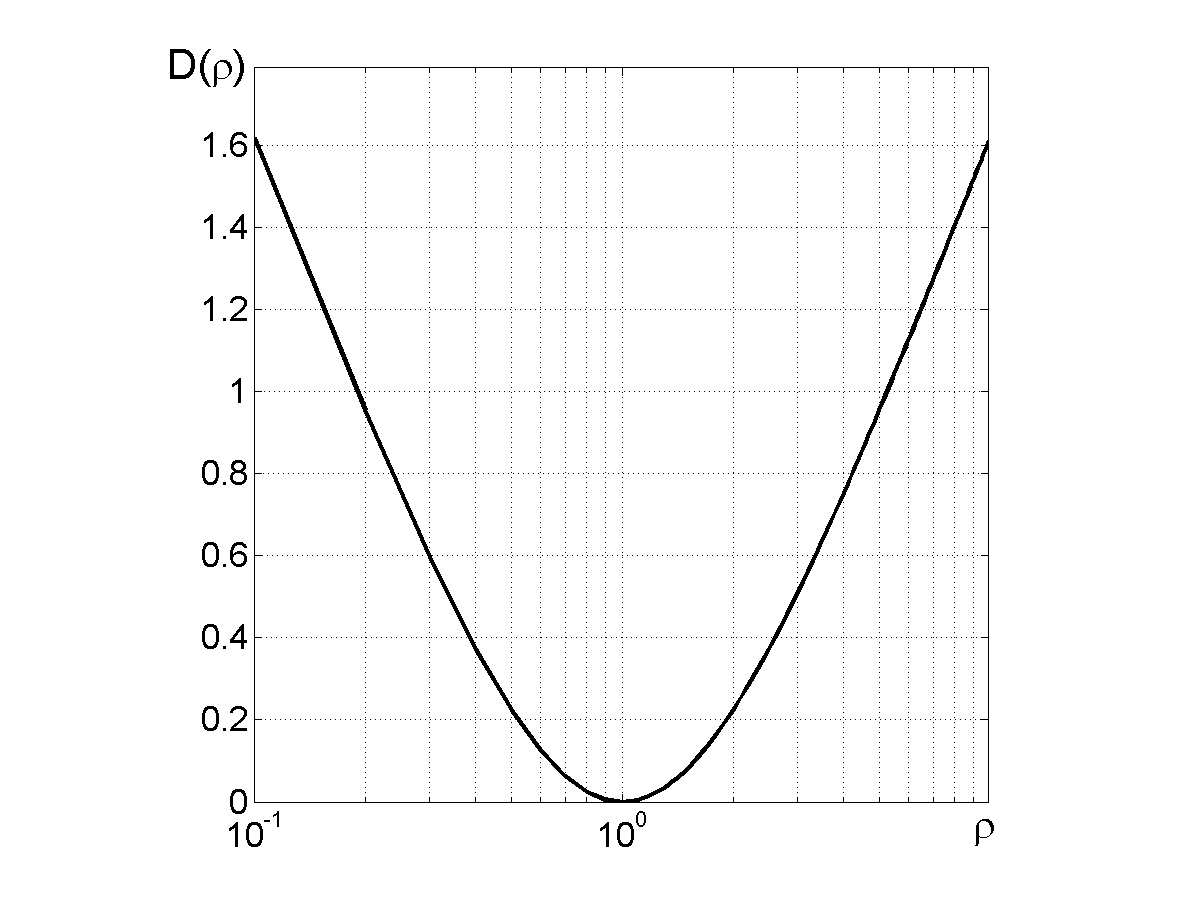} &
        \includegraphics[trim={2.4cm 0 2.4cm 0},clip,width=4.0cm]{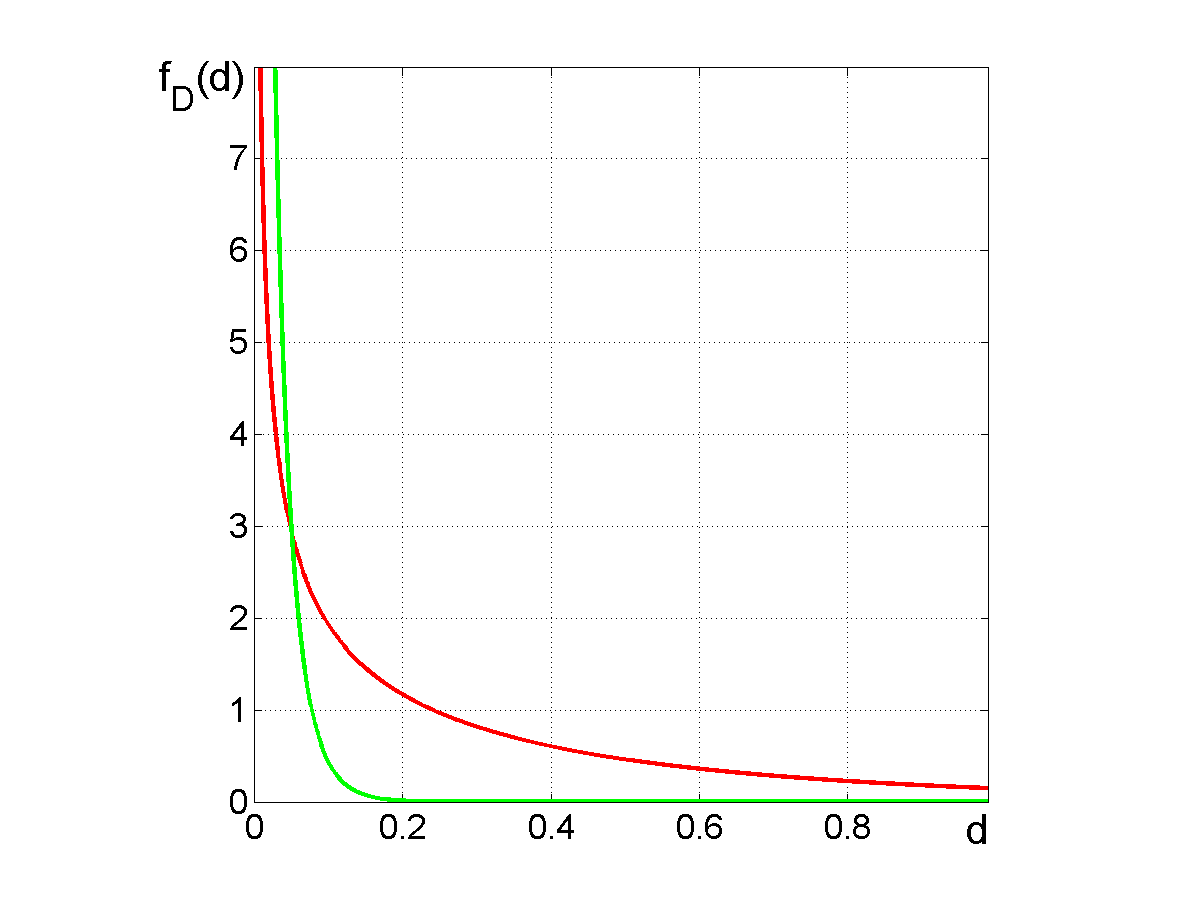}
    \end{tabular}
    \caption{
Left:  plot of the speckle-free distance $D(\rho)=\log(\rho/2+1/2\rho)$ in semilog axes.
Right: theoretical pdf of $D$ for equal signal intensity ($\rho_x=1$) and unit-mean Gamma-distributed speckle, $L$=1 (red) and $L$=16 (green).}
\label{fig:about_D}
\end{figure}

In the first case, $u(s)=u(t)=1$,
the distance depends only on the ratio of signal intensities $\rho_x^2=x(s)/x(t)$, that is
\begin{eqnarray}
    D_{\rm SF}[z(s),z(t)] = \log \!\left[ \frac{x(s)+x(t)}{2\sqrt{x(s)\,x(t)}} \right] = \log \!\left[ \frac{\rho_x}{2}+\frac{1}{2\rho_x} \right]
\end{eqnarray}
In Fig.\ref{fig:about_D}(left) we plot $D_{\rm SF}$ as a function of the signal intensity ratio.
For $\rho_x$ close to 1, the distance is rather flat around the minimum, zero, and begins growing linearly (in semilog axes) only for much larger/smaller values.
Therefore, it is not much discriminative for samples with relatively close intensity.

On the other hand, with homogeneous signal, $x(s)=x(t)$, the distance is the random variable
\be
    D[u(s),u(t)] = \log \left[ \frac{u(s)+u(t)}{2\sqrt{u(s)u(t)}} \right]
\ee
with $u(s)$ and $u(t)$ independent Gamma distributed RV's, with unit mean, and shape parameter equal to the number of looks of the image, $L$.
In Appendix A we compute the probability density function (pdf) of $D$ as a function of $L$.
Fig.\ref{fig:about_D}(right) shows two such pdf's, for $L$=1 and $L$=16.
In the relatively uninteresting case of $L$=16 (low noise) the pdf is highly peaked around 0, that is, homogeneous pixels do have small SAR-domain distances.
However, in the more interesting and relevant case of $L=1$ (single-look images) the pdf is much flatter, and has a long non-negligible tail.

The plots of Fig.\ref{fig:about_D} can be used to gain insight into the discriminative power of the SAR-domain distance.
As an example, for $x(s)/x(t)=\rho_x=2$, the speckle-free distance is about 0.2.
This should allow one to recognize that these pixels are non-homogeneous.
However, for homogeneous pixels, $x(s)=x(t)$, the single-look distance exceeds 0.2 with probability 0.4.
This means that a 0.2 distance provides little or no information on the quality of the predictor.

\begin{figure}[t]
\centering
    \begin{tabular}{cc}
    \setlength{\tabcolsep}{1pt}
        \includegraphics[trim={2.4cm 0 2.4cm 0},clip,width=4.0cm]{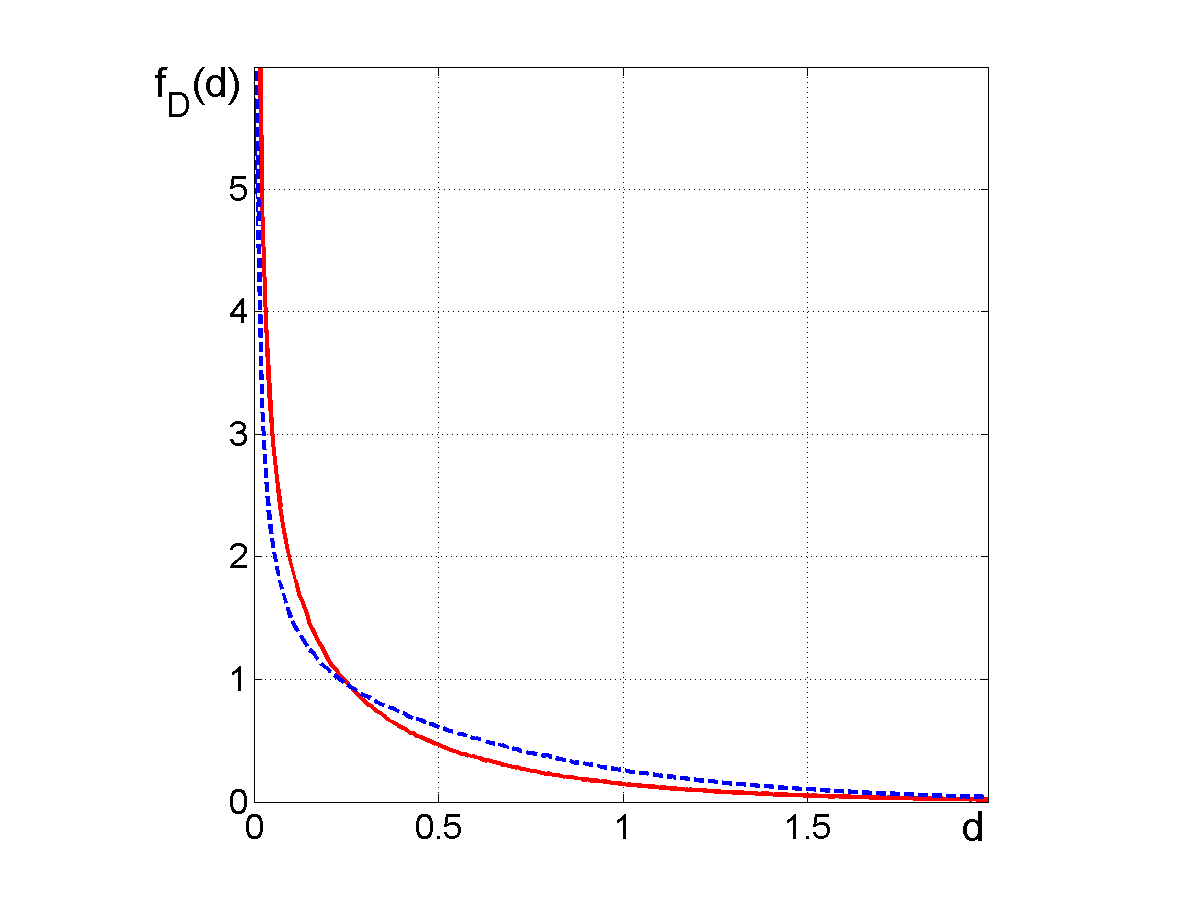} &
        \includegraphics[trim={2.4cm 0 2.4cm 0},clip,width=4.0cm]{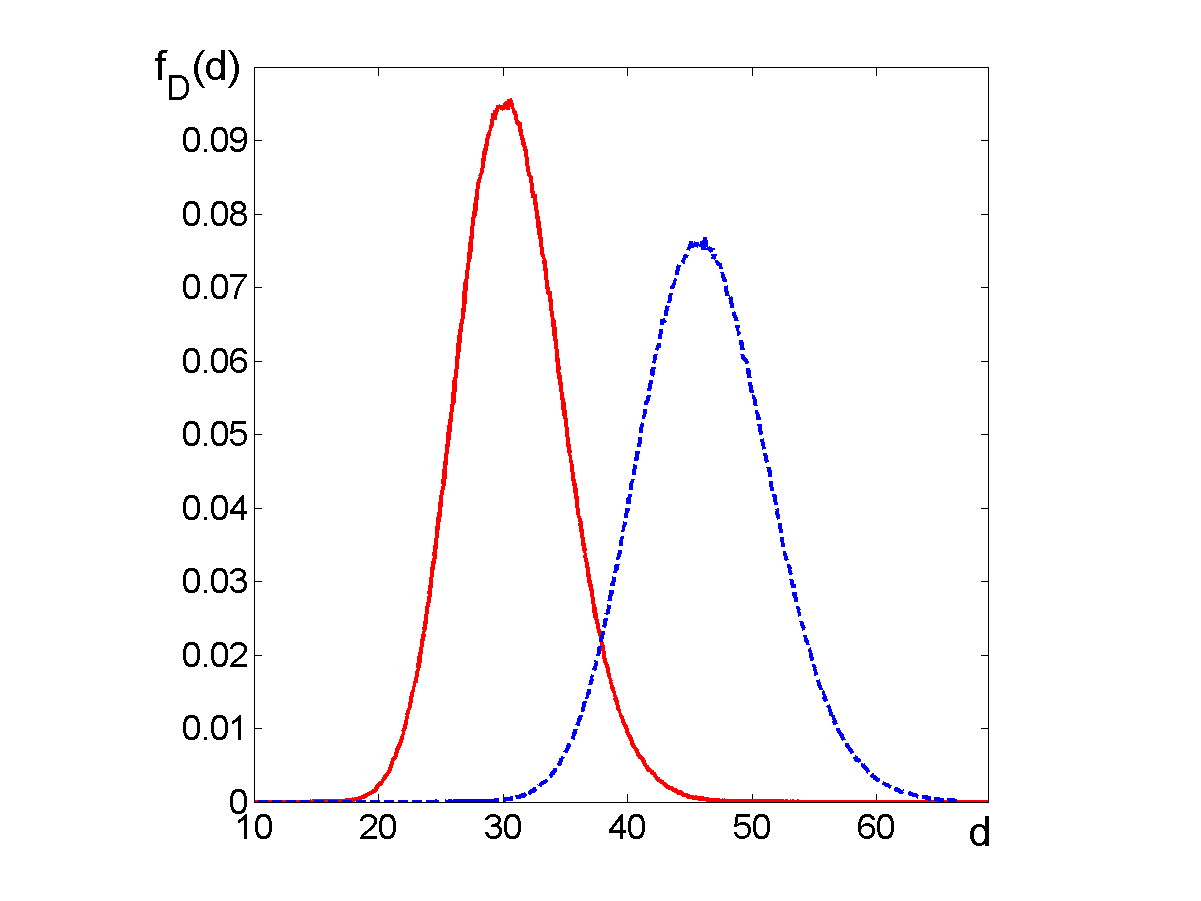}
    \end{tabular}
    \caption{Empirical pdf of D with unit-mean Gamma-distributed speckle, $L$=1, for $\rho_x=1$ (solid red) and $\rho_x=2$ (dashed blue).
             Left: pixel-wise distance. Right: patch-wise distance, with 10$\times$10-pixel patches.}
\label{fig:MonteCarlo_pdf}
\end{figure}

For a more precise analysis, the pdf of the distance for arbitrary signal intensity ratio is necessary.
Lacking the closed-form pdf, we resort to MonteCarlo simulation.
Fig.\ref{fig:MonteCarlo_pdf}(left) shows the empirical pdf's for $\rho_x^2=1$ (solid red) and $\rho_x^2=2$ (dashed blue).
As expected, they largely overlap, indicating that no reliable discrimination is possible,
and justifying, in hindsight, the classification-based solution proposed in \cite{Verdoliva2015}.
In this work, however, we use patch-wise distances obtained by summing pixel-wise distances over many samples.
Assuming, for the sake of simplicity, two patches with constant signal intensity ratio, that is
\be
    x(s+k)/x(t+k) = \rho_x^2, \ls \forall k \in \mathbb{P}
\ee
the patch-wise distance becomes the sum of $N$ independent identically distributed RV's, well approximated by a Gaussian law.
Fig.\ref{fig:MonteCarlo_pdf}(right) shows again the estimated pdf's for $\rho_x^2=1$ and $\rho_x^2=2$ when 10$\times$10-pixel patches are considered.
As expected, same-signal and different-signal patches have now well separated pdf's,
suggesting that a test based on patch-wise distances can provide reliable indications.

Note that, for good predictors, that is, patches similar to the target, the constant-ratio hypothesis with $\rho_x=1$ is quite reasonable.
Under this hypothesis, mean and variance of the pixel-wise distance, $\mu_D$ and $\sigma_D^2$, are computed in Appendix A,
and hence the approximating Gaussian curve is perfectly known.
Therefore, we can set the threshold test with a Neyman-Pearson criterion,
deciding in advance which fraction of the good predictors can be lost to ensure rejection of virtually all the bad ones.

\section{Exploring key parameters}

\begin{figure*}
	\centering\footnotesize
	\begin{tabular}{c@{\hspace{1mm}}c@{\hspace{1mm}}c@{\hspace{1mm}}c@{\hspace{1mm}}c@{\hspace{1mm}}c}
		\includegraphics[width=2.8cm]{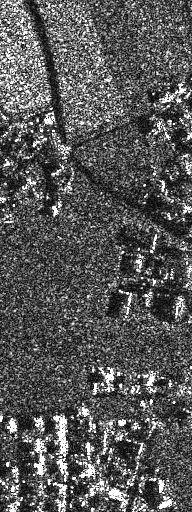} &
		\includegraphics[width=2.8cm]{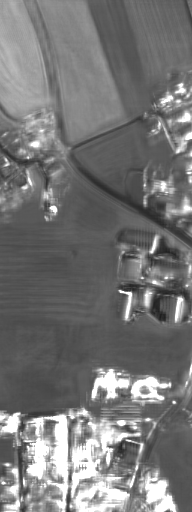} &
		\includegraphics[width=2.8cm]{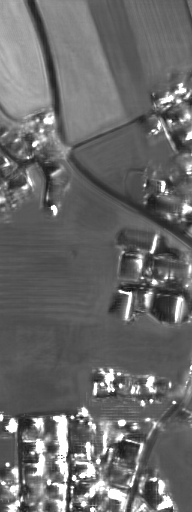} &
		\includegraphics[width=2.8cm]{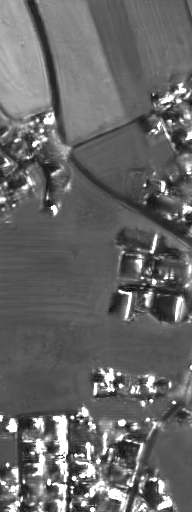} &
		\includegraphics[width=2.8cm]{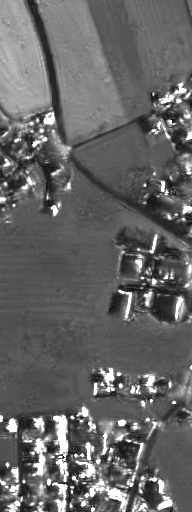} &
		\includegraphics[width=2.8cm]{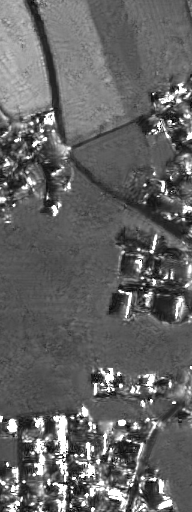} \\
        \rule{0mm}{3mm} original & $T=\infty$ & $T=1+4\sigma$ & $T=1+2\sigma$ & $T=1+\sigma$ & $T=1$
    \end{tabular}
    \caption{Visual quality of filtered images as a function of the SAR-distance threshold.
    When all patches contribute to the estimate ($T=\infty$) large filtering artifacts appear.
    A very small threshold ($T=1$), instead, causes the rejection of too many predictors, reducing the speckle rejection ability of the filter.}
    \label{fig:explore_threshold}
\end{figure*}

Like all numerical algorithms,
the proposed method depends on several key parameters which impact significantly on performance.
Some of them are related to nonlocal means and are set based on literature results, like search area, 39$\times$39, and patch size, 8$\times$8.
Others are set based on preliminary experiments to meet all contrasting quality requirements.
Of these latter, decay and balance parameters, $\lambda$ and $\gamma$, have a rather obvious meaning and need no special analysis.
Instead, it is worth gaining more insight into how the test threshold, $T$, and the maximum number of predictors, $S_0$, impact on performance.
To this end we carry out a visual analysis on the T1 clip, shown in Fig.\ref{fig:T-SAR-data},
which displays both agricultural and urban areas, and hence allows for a study of all features of interest.

\subsection{Threshold $T$}

Fig.\ref{fig:explore_threshold} refers to a 512$\times$192 strip of the T1 image, thin enough to allow for a simple visual comparison of results.
From left to right we show the original SAR strip,
and 5 filtered versions obtained with threshold $T$ in $\{\infty, 1 \plus 4\sigma, 1 \plus 2\sigma, 1 \plus \sigma, 1\}$,
where $\sigma=\sigma_P$ is the standard deviation of the normalized SAR distance for homogeneous signal.
With $T=\infty$ the test does not operate, and bad predictors contribute to the estimate, causing a severe impairment of the filtered image.
Major artifacts are visible in urban areas, due to strong reflectors, but problems arise also in other areas, for example the dark road in the top almost vanishes with filtering.
The test on SAR distances solves most of these problems.
Even a large threshold, $T=1+4\sigma$, which excludes only a tiny fraction of good predictors, removes most bad ones.
Lowering the threshold to $T=1+2\sigma$, a more selective test is obtained, and a further quality improvement is observed.
With further smaller values, however, a large part of good predictors is rejected together with the bad ones,
reducing the speckle rejection ability of the filter in homogeneous areas, and causing the appearance of residual noise.

\begin{figure}[!b]
	\centering\footnotesize
	\begin{tabular}{c@{\hspace{1mm}}c}
        \includegraphics[width=4.0cm]{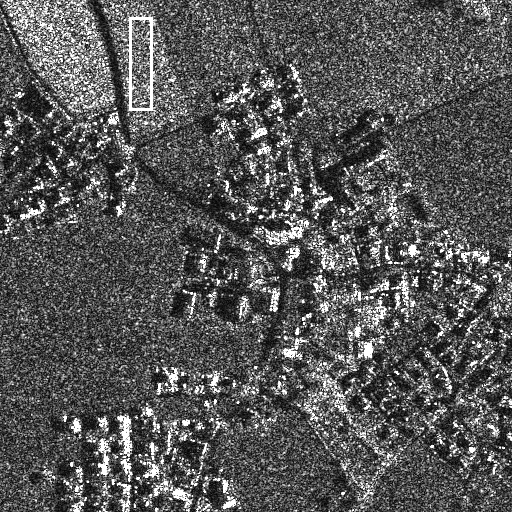} &
        \includegraphics[width=4.0cm]{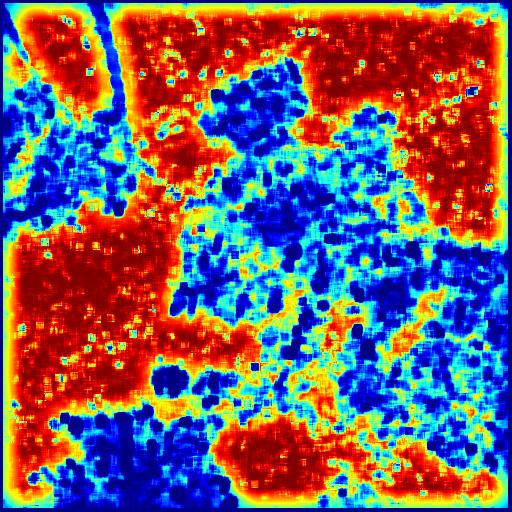}
    \end{tabular}
    \caption{False-color representation of the number of predictor patches passing the test at level $T=1+2\sigma$, from 0\% (dark blue) to 100\% (intense red).}
    \label{fig:heatmap}
\end{figure}

Fig.\ref{fig:heatmap} provides further insight into how the test impacts on the number of predictors.
Besides the original T1 clip, on the left, we show a false-color map of the number of patches that pass the test with the selected threshold $T=1+2\sigma$,
going from 1 (dark blue), to the maximum $S$=39$\times$39 (intense red).
It clearly appears that in urban areas only a few patches survive the test, those that are structurally similar to the target, thus avoiding filtering artifacts.
Instead, a very large number of patches pass the test in homogeneous regions, ensuring good speckle rejection.

\subsection{Maximum number of predictors $S_0$}

\begin{figure}
	\centering\footnotesize
	\begin{tabular}{c@{\hspace{1mm}}c}
        \rotatebox{90}{~~~~~~guide}       & \includegraphics[width=8.0cm]{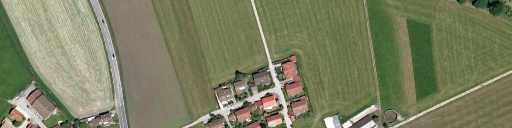} \\
        \rotatebox{90}{~~~~~original}     & \includegraphics[width=8.0cm]{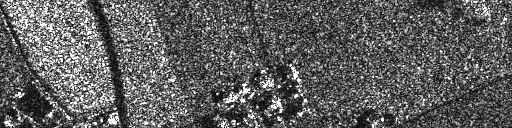} \\
        \rotatebox{90}{~~~~~$S_0=S$}      & \includegraphics[width=8.0cm]{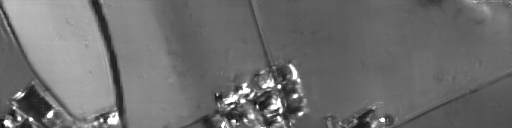} \\
        \rotatebox{90}{~~~~$S_0\!=\!256$} & \includegraphics[width=8.0cm]{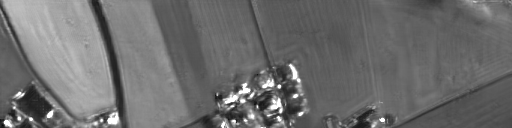} \\
        \rotatebox{90}{~~~~~$S_0=64$}     & \includegraphics[width=8.0cm]{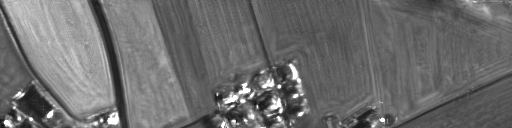}
    \end{tabular}
    \caption{Visual quality of filtered images as a function of the maximum number of predictors in the search area.
    In homogeneous areas, using all predictors, $S_0=S$, causes oversmoothing. With too few predictors, $S_0=64$, speckle rejection is less effective. }
    \label{fig:explore_number}
\end{figure}

To visualize the impact of the maximum number of predictors, $S_0$, on filtering quality, we use a thin horizontal strip of the T1 image
which contains mostly a mosaic of homogeneous regions and some roads.
In fact, in urban areas, the number of predictors is already limited by the SAR-domain test and no further constraint is needed.
Fig.\ref{fig:explore_number} shows, from top to bottom,
the optical guide, the original SAR data, and the output of the filter with $S_0$=1521 (all the patches), 256, and 64.
In the first case, the filtered image appears oversmoothed.
For example, the thin white road on the right is lost, and the boundary between the fields on the left is much smeared.
These structures and others are recovered in the second case, $S_0$=256.
On the down side, some textures appear in the fields which cannot be spotted in the original SAR image, and may raise the suspect of incorrect behavior.
However, one must remember that only original SAR data are averaged.
The optical guide can only give priority to some patches over others and, so to speak, ``combing'' the data in a certain direction, but the data are all SAR.
This is confirmed by the thin white road on the right.
A careful inspection reveals clear traces of the road in the original data, which are reinforced by the guided filtering.
With these considerations, also the third case, with $S_0$=64, is probably acceptable,
however we prefer to ensure a stronger speckle rejection and a smoother output and use $S_0$=256 as default parameter.

Based on these experiments,
we select eventually a configuration with parameters $T=1+2\sigma=1.34, S_0=256, \lambda=0.002$ and $\gamma=0.15$, ensuring sharp details and good speckle rejection.
However, we also consider a more conservative configuration, with $S_0=S$ and $\lambda=0.004$, which outputs somewhat smoother images.
Of course, for other datasets, these parameters may require some fine tuning, also due to different dynamics.
With our COSMO dataset, for example, we only needed to multiply all $\lambda$'s by a factor 4.

\section{Experimental analysis}

In this Section, we discuss the results of several experiments on two real-world datasets,
where the proposed method is compared with several state-of-the-art references.
In the following, we describe datasets, quality assessment criteria, reference methods and, finally, numerical and visual results.

\subsection{Datasets}

We designed two SAR/optical datasets, called for short T-SAR and COSMO, in the following.


\begin{figure}
	\centering\footnotesize
	\begin{tabular}{c@{\hspace{1mm}}c@{\hspace{1mm}}c@{\hspace{1mm}}c}
		\includegraphics[width=2.1cm]{./Figures/Datasets/T1_SAR.png} &
		\includegraphics[width=2.1cm]{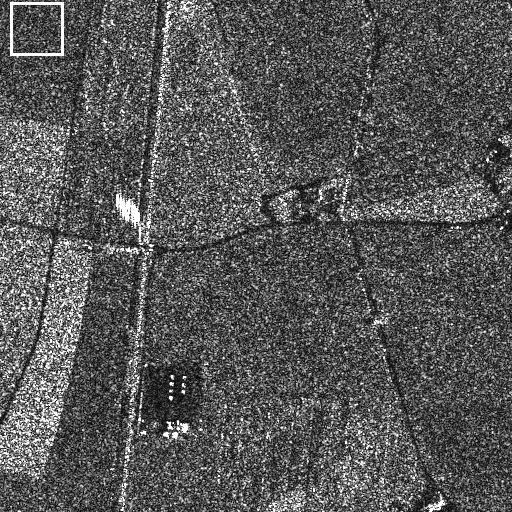} &
		\includegraphics[width=2.1cm]{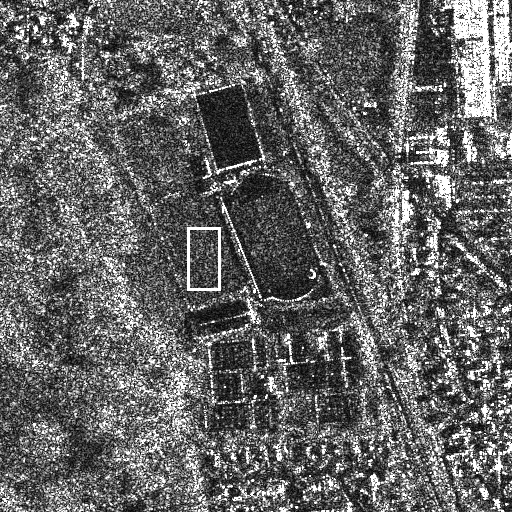} &
		\includegraphics[width=2.1cm]{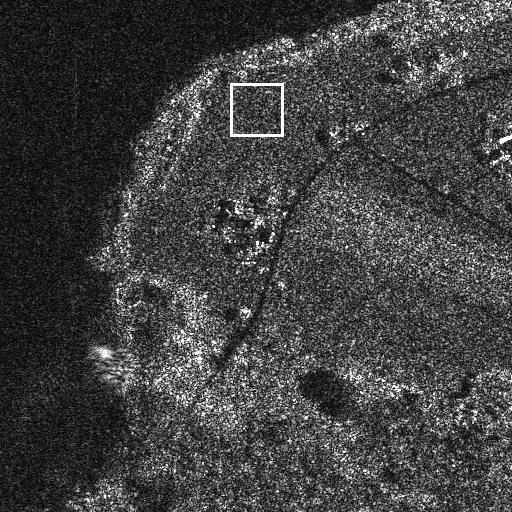} \\
        \vspace{-3mm} & & & \\
		\includegraphics[width=2.1cm]{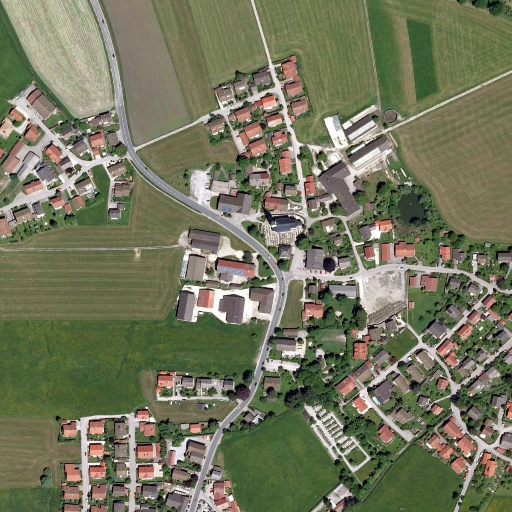} &
		\includegraphics[width=2.1cm]{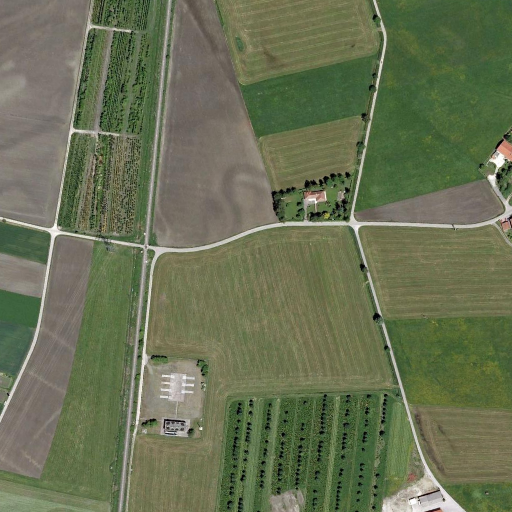} &
		\includegraphics[width=2.1cm]{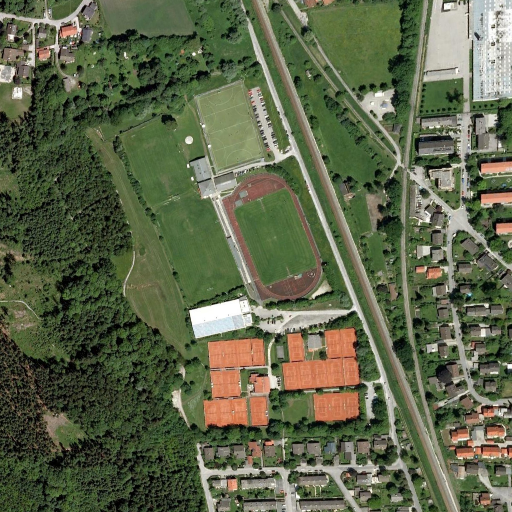} &
		\includegraphics[width=2.1cm]{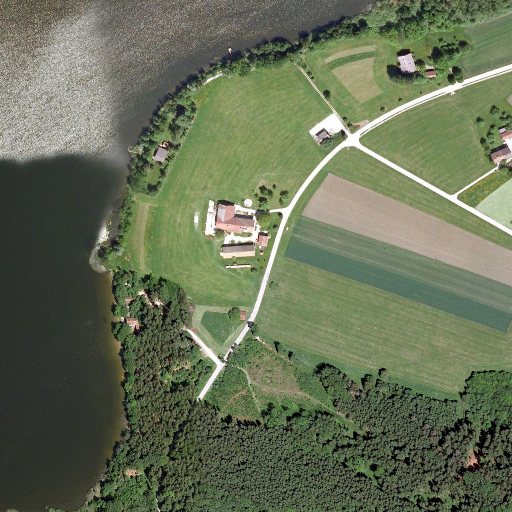} \\
    \end{tabular}
    \caption{SAR-optical pairs of the T-SAR dataset: T1...T4 from left to right.}
    \label{fig:T-SAR-data}
\end{figure}

\begin{figure}
	\centering\footnotesize
	\begin{tabular}{c@{\hspace{1mm}}c@{\hspace{1mm}}c@{\hspace{1mm}}c@{\hspace{1mm}}c@{\hspace{1mm}}c@{\hspace{1mm}}c@{\hspace{1mm}}c}
		\includegraphics[width=2.1cm]{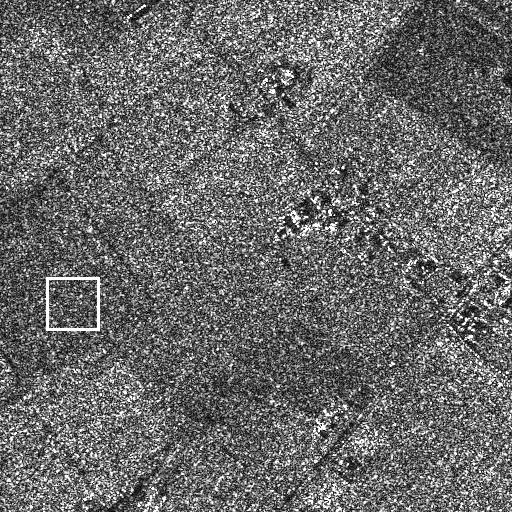} &
		\includegraphics[width=2.1cm]{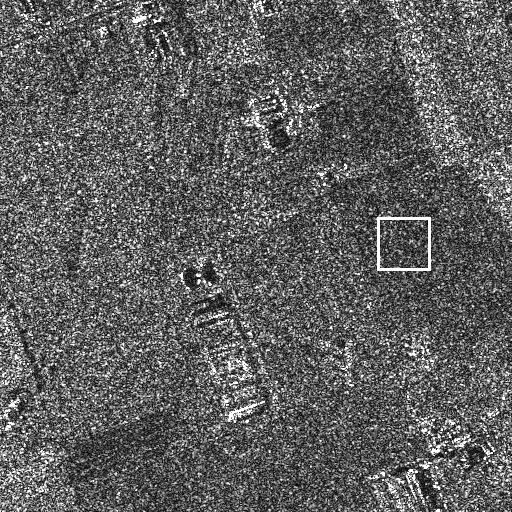} &
		\includegraphics[width=2.1cm]{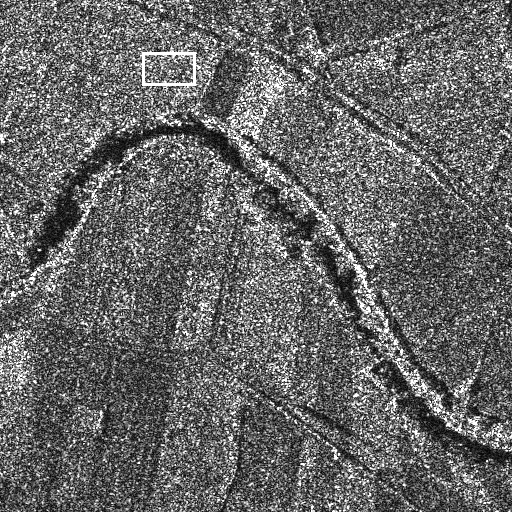} &
		\includegraphics[width=2.1cm]{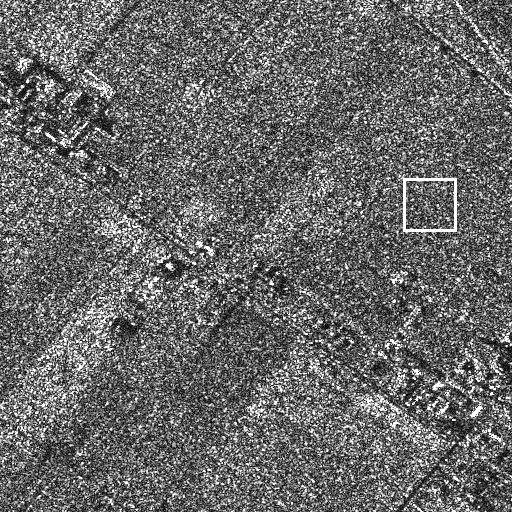} \\
        \vspace{-3mm} & & & & & & & \\
		\includegraphics[width=2.1cm]{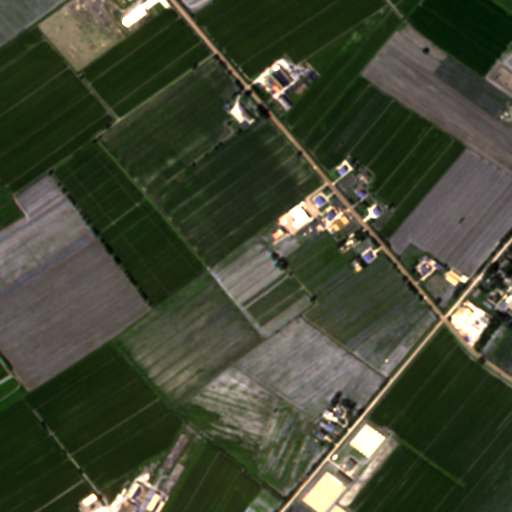} &
		\includegraphics[width=2.1cm]{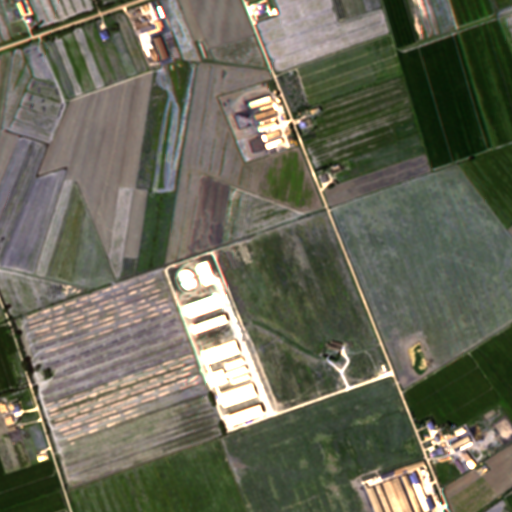} &
		\includegraphics[width=2.1cm]{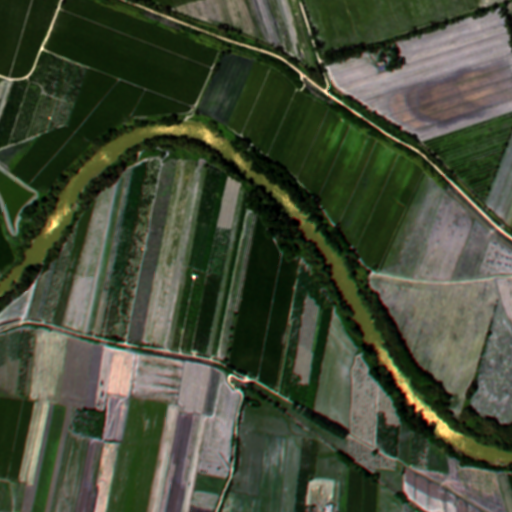} &
		\includegraphics[width=2.1cm]{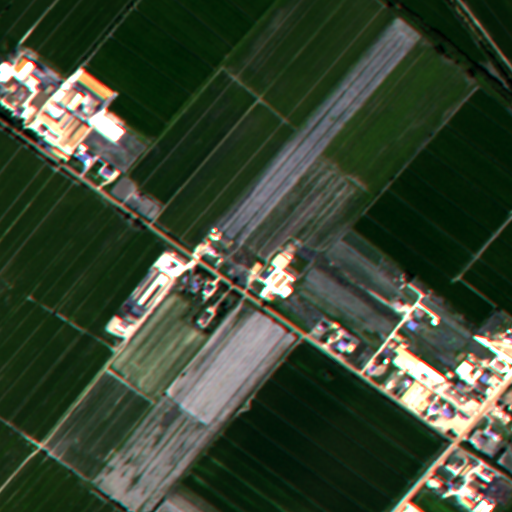} \\
    \end{tabular}
    \caption{SAR-optical pairs of the COSMO dataset: C1...C4 from left to right.}
    \label{fig:COSMO-data}
\end{figure}

The T-SAR dataset includes four 512$\times$512-pixel clips
extracted from a large single-look TerraSAR-X image (courtesy of \copyright Infoterra GmbH) acquired over Rosenheim (D) in spotlight mode with single HH polarization on January 27th, 2008.
For this image, we do not have an optical reference of comparably high quality,
therefore we resort to the freely available RGB optical images provided by Google Earth Pro.
This ''barebone'' setting is of particular interest for us, since we want the proposed method to be adopted with minimal effort also by users with limited budget.
In the Google Earth repository, we found, as of october 2018, several images of the Rosenheim region,
spanning from 2002 to 2017, with the closest one acquired on December 31st, 2009, about two years after the target.

In Fig.\ref{fig:T-SAR-data} we show the four SAR-optical pairs, called T1, T2, T3, and T4, from now on.
Each optical image was co-registered with the corresponding SAR image, used as master.
The available geocoding information was used for a first raw alignment, refined manually based on a few prominent keypoints.
On the average, the co-registration process took about 5 minutes per image.
Despite the large temporal gap, all pairs match quite well.
On the other hand, some mismatches in the test set are welcome, because good optical references may be unavailable for several reasons, like the presence of clouds,
and the proposed method must provide sensible results also in the presence of mismatches or missing data.

The four SAR-optical pairs of the T-SAR dataset are available online, at www.grip.unina.it, to allow other researchers to experiment with the very same data.
Moreover, to ensure the reproducibility of our research, the executable code of the proposed method is also available online at the same address.

The COSMO dataset includes four 512$\times$512-pixel clips
extracted from a large COSMO-SkyMed image acquired over the province of Caserta (I) on July 18th, 2011.
In this case, as optical guide we can rely on a GeoEye-1 multispectral image, acquired over the same region on July 26th, 2010
(both images curtesy of the Italian Aerospace Research Center).
The same co-registration process as for T-SAR was used.
In Fig.\ref{fig:COSMO-data} we show the four SAR-optical pairs, called C1, C2, C3, and C4, from now on, with a suitable RGB rendering of the 4-band Geoeye-1 images.

\subsection{Quality assessment criteria}

Despite intense research on SAR despeckling, quality assessment is still an open problem \cite{DiMartino2014}.
A good filter should guarantee both effective suppression of speckle and faithful preservation of informative image details, such as region boundaries or man-made structures.
These are contrasting requirements,
as intense image smoothing, which ensures speckle removal, tends to smear all relevant high-frequency details.
Therefore, these two aspects should be analyzed individually.

To this end, we consider here two objective indicators,
{\it  i)} the equivalent number of looks (ENL), and
{\it ii)} the ratio image structuredness (RIS).
In any case, we leave the last word to the visual inspection of filtered and ratio images.

The ENL is the squared ratio between the mean and the standard deviation of the signal computed over a homogeneous region of the image.
In Fig.\ref{fig:T-SAR-data} and Fig.\ref{fig:COSMO-data} the regions used to compute the ENL are shown enclosed in a white box.
Before filtering, the ENL approximates the number of looks of the SAR image.
After despeckling, instead, it should be as large as possible, as the filtered image is supposed to approach a constant.

The RIS is computed on the ratio image, that is, the ratio between original and filtered images.
With ideal filtering and fully developed speckle, the ratio image becomes a field of i.i.d. speckle samples.
Imperfect filtering, instead, causes the leakage of image structures in the ratio, which looses its i.i.d. nature.
Hence, neighboring pixels tend to be more similar to one another.
We measure this tendency through a suitable function of their joint pdf
\begin{equation}
    p(i,j) = \Pr(r(t)=i,r(s)=j)
\end{equation}
with $r$ the quantized ratio image, and $t,s$ two 4-connected sites.
Inspired by Gomez {\it et al.} \cite{Gomez2017} we use the homogeneity textural descriptor proposed by Haralick \cite{Haralick1973}
\begin{equation}
    H = \sum_i\sum_j p(i,j) \frac{1}{(i-j)^2-1}
\end{equation}
and compare it with the reference value, $H_0$, computed on the product of the marginals, $p_0(i,j)=p(i)p(j)$,
obtaining eventually the RIS index defined as
\begin{equation}
    {\rm RIS} = 100 \times \frac{H-H_0}{H_0}
\end{equation}

\subsection{Reference methods}

We compare the proposed method with a few selected reference methods, chosen because of their diffusion in the community and good performance.
More precisely, we include
\begin{itemize}
\item   Enhanced-Lee \cite{Lopes1990}: an enhanced version of Lee's adaptive local filter \cite{Lee1980}, widespread in the community;
\item   PPB \cite{Deledalle2009}: a patch-based iterative nonlocal filter, where the output is given by a weighted maximum likelihood estimator with data-driven weights;
\item   SAR-BM3D \cite{Parrilli2012}: the SAR-domain adaptation of the nonlocal BM3D filter \cite{Dabov2007}, with wavelet shrinkage and Wiener filtering;
\item   FANS \cite{Cozzolino2014}: a faster and spatially-adaptive version of SAR-BM3D, which ensures a better speckle rejection in homogeneous areas;
\item   G-BF \cite{Verdoliva2015}: our previous optical-guided pixel-wise despeckling method, based on generalized bilateral filter.
\end{itemize}
PPB and SAR-BM3D, in particular, represent sort of two limiting cases,
with PPB ensuring very strong speckle rejection in homogeneous areas at the cost of some smearing of high-frequency details,
and SAR-BM3D much better at preserving details but less effective otherwise.

For all methods, we selected parameters as suggested in the original papers or, lacking clear indications, such to optimize filtering quality.
For the proposed method, we consider the two configurations described at the end of Section IV,
resulting in two versions, a first one (sharp) which makes a more aggressive use of the optical guide, and a second one (smooth) more conservative.

\subsection{Results}

\renewcommand{\arraystretch}{1.1}
\renewcommand{\ru}{\rule{0mm}{3.2mm}}
\renewcommand{\i}[1]{{\it #1}}
\begin{table}
{\footnotesize
\caption{Equivalent number of looks (ENL) on T-SAR clips.}
\centering
\setlength{\tabcolsep}{4pt}
\begin{tabular}{c|rrrrrrr}
\ru        Clip &   \local &      \ppb~ &     \bmtd &     \fans &      \gbf &    \gnlmA  &    \gnlmC  \\ \hline
\ru          T1 &   11.4~  &    326.3~  &     9.1~  &    77.8~  &   140.5~  &    223.5~  &    763.7~  \\
\ru          T2 &    9.3~  &    363.3~  &     8.6~  &    40.0~  &   162.1~  &    235.9~  &    679.4~  \\
\ru          T3 &    9.6~  &    242.1~  &     6.0~  &    36.3~  &    55.7~  &    124.3~  &    409.1~  \\
\ru          T4 &    7.1~  &    194.9~  &     5.0~  &    22.9~  &    12.0~  &    135.2~  &    552.3~  \\ \hline
\ru \i{average} & \i{9.3}~ & \i{281.6}~ &  \i{7.2}~ & \i{44.2}~ & \i{92.6}~ & \i{179.7}~ & \i{601.1}~ \\
\end{tabular}                                                                                                                             %
\label{tab:ENL_T}
}
\end{table}

\begin{table}
{\footnotesize
\caption{Equivalent number of looks (ENL) on COSMO clips.}
\centering
\setlength{\tabcolsep}{4pt}
\begin{tabular}{c|rrrrrrr}
\ru        Clip &   \local &     \ppb~  &    \bmtd &     \fans &       \gbf &    \gnlmA  &    \gnlmC  \\ \hline
\ru          C1 &    8.0~  &    236.0~  &    4.6~  &    27.8~  &    203.8~  &    199.6~  &    751.3~  \\
\ru          C2 &    6.4~  &    159.1~  &    3.5~  &    15.4~  &     64.8~  &     72.9~  &    248.0~  \\
\ru          C3 &    9.2~  &    281.7~  &    5.8~  &    33.8~  &    107.7~  &    157.7~  &   1208.1~  \\
\ru          C4 &    8.0~  &    315.1~  &    5.2~  &    39.3~  &     70.2~  &    116.8~  &    442.5~  \\ \hline
\ru \i{average} & \i{7.9}~ & \i{248.0}~ & \i{4.8}~ & \i{29.1}~ & \i{111.6}~ & \i{136.7}~ & \i{662.5}~ \\
\end{tabular}                                                                                                                             %
\label{tab:ENL_C}
}
\end{table}

Tab.\ref{tab:ENL_T} and Tab.\ref{tab:ENL_C} report the ENL results for all T-SAR and COSMO clips, respectively, with the average values in the last row.
Despite the obvious variations from clip to clip,
clear indications emerge from these data, well summarized by the average values, which are almost identical for the two datasets.
The proposed method ensures a very strong rejection of speckle, with average ENL beyond 100 for the first version and over 600 for the second one.
Among reference methods, only PPB provides comparable results, with ENL around 250, while enhanced Lee and BM3D even remain below 10.
Our previous optical-guided filter also provides a good ENL, about 100 on the average.

\begin{figure}
	\centering\footnotesize
	\begin{tabular}{c@{\hspace{1mm}}c@{\hspace{1mm}}c}
        enhanced-Lee & PPB & SAR-BM3D \\
		\includegraphics[width=2.8cm]{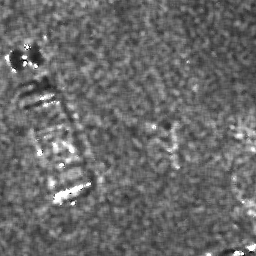} &
		\includegraphics[width=2.8cm]{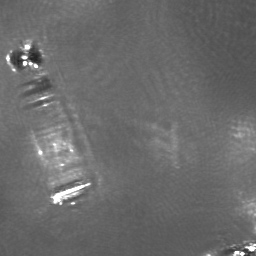} &
		\includegraphics[width=2.8cm]{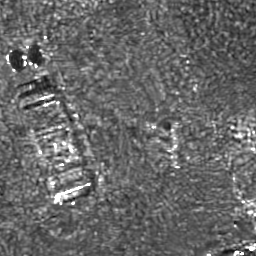} \vspace{1mm} \\
        optical guide & single-look SAR data & FANS \\
		\includegraphics[width=2.8cm]{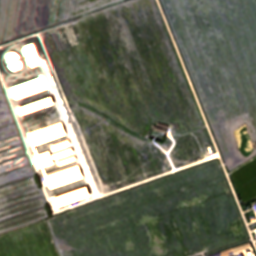} &
		\includegraphics[width=2.8cm]{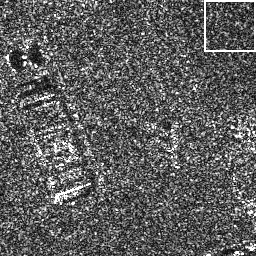} &
		\includegraphics[width=2.8cm]{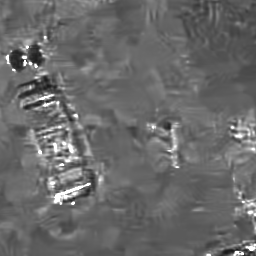} \vspace{1mm} \\
        G-BF & G-NLM sharp & G-NLM smooth \\
		\includegraphics[width=2.8cm]{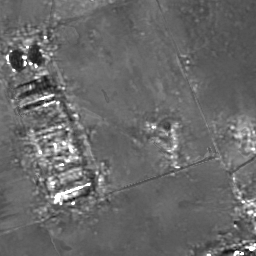} &
		\includegraphics[width=2.8cm]{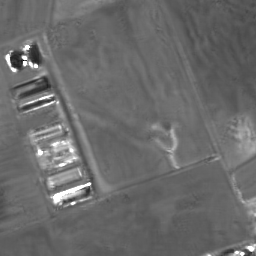} &
		\includegraphics[width=2.8cm]{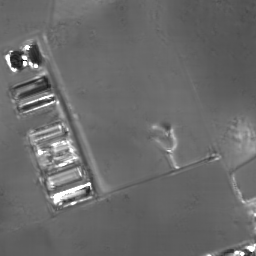} \vspace{1mm} \\
    \end{tabular}
    \caption{Filtering results for the C2 clip. The single-look original is shown in the center for easy comparison.
    Reference methods all present some shortcomings: limited speckle rejection, loss of resolution, or filtering artifacts.
    Thanks to the optical guide, G-NLM ensures a much better image quality.}
    \label{fig:comparison_C2}
\end{figure}

As already said, however, these results may be misleading if not accompanied by visual analysis.
Therefore, we now show and comment the output of all filters for some relevant details selected from our datasets.
To allow the reader to conduct a more thorough inspection, we publish online, at www.grip.unina.it, the results of all methods under comparison on all clips of our datasets.
Fig.\ref{fig:comparison_C2} shows a 256$\times$256 section of the C2 clip.
Except for some buildings in the left part, the scene includes only fields and some thin roads; the region used to compute the ENL is in the upper-right corner.
The visual inspection reveals a number of phenomena hardly captured by the ENL or other numerical measures.
It confirms the limited speckle suppression of enhanced Lee and SAR-BM3D, but also the well-known detail preservation ability of the latter.
FANS produces a smoother output, but many wavelet-related artifacts appear, which impair significantly the perceived quality.
Likewise, PPB is very effective in removing speckle, but introduces ``brushstroke'' patterns and, what is worse, smears edges and buildings.
Also our previous optical-guided filter generates some artifacts in smooth areas, and produces annoying halos of residual speckle in high-activity areas.
The proposed optical-guided nonlocal means, in both versions, produces images of much better quality.
While man-made structures are faithfully preserved (compare with the original SAR image) speckle is largely rejected in all homogeneous areas.
In the ``smooth'' version, these areas become basically flat,
while in the ``sharp'' version some subtle patterns emerge as a result of the SAR data ``combing''.
It remains to understand whether these are real structures, hidden in the SAR data and recovered through filtering, or else they come from the alien optical guide.
Lacking a clean reference, no ultimate answer can be given.
Nonetheless, some of these structures can be clearly spotted in the original SAR image, though overwhelmed by speckle.
For example, the diagonal dark strip in the center, or the parallel thin strips above the central bright fork, which in fact are both captured also by enhanced Lee and SAR-BM3D.
Therefore, a reasonable guideline could be to use the sharp version if the optical guide is temporally close to the SAR image, and the smooth version otherwise.
In any case, it is clear that both versions of G-NLM work much better than G-BF, based itself on the use of an optical guide,
since they ensure a much better suppression of speckle and do not introduce filtering artifacts.

\begin{figure}
	\centering\footnotesize
	\begin{tabular}{c@{\hspace{1mm}}c}
        optical guide & single-look SAR data \\
		\includegraphics[width=4.2cm]{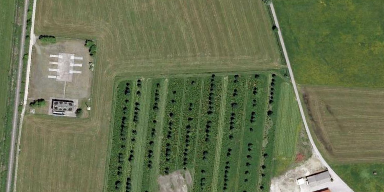} &
		\includegraphics[width=4.2cm]{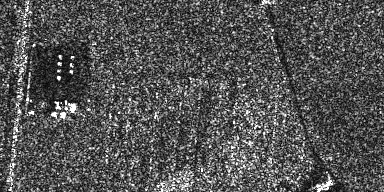} \vspace{1mm} \\
        enhanced-Lee & PPB \\
		\includegraphics[width=4.2cm]{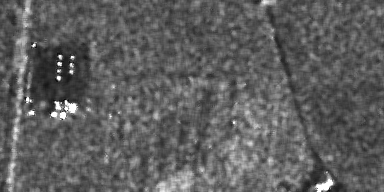} &
		\includegraphics[width=4.2cm]{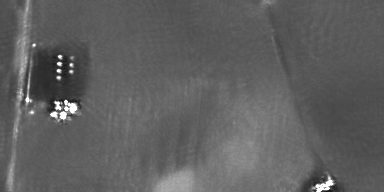} \vspace{1mm} \\
         SAR-BM3D & G-BF \\
		\includegraphics[width=4.2cm]{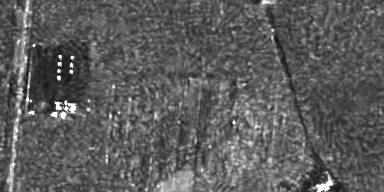} &
		\includegraphics[width=4.2cm]{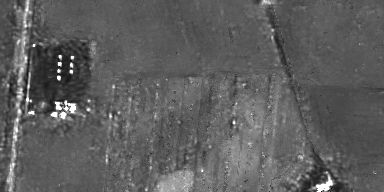} \vspace{1mm} \\
         G-NLM sharp & G-NLM smooth \\
		\includegraphics[width=4.2cm]{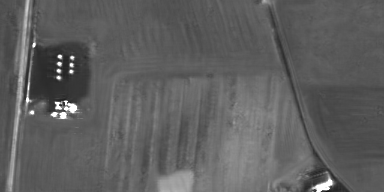} &
		\includegraphics[width=4.2cm]{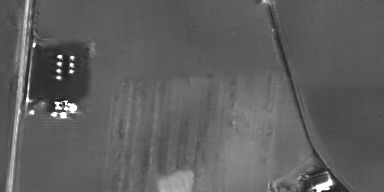} \vspace{1mm} \\
    \end{tabular}
    \caption{Filtering results for the T2 clip.}
    \label{fig:comparison_T2}
\end{figure}

All the above consideration are reinforced by further visual analyses.
Here we only show another detail, in Fig.\ref{fig:comparison_T2}, a 192$\times$384 section from the bottom of the T2 clip.
Again, the proposed method ensures a better speckle rejection (especially the smooth version) and detail preservation (especially the sharp version) than all references,
with an excellent overall quality, considering also the very noisy single-look original data.
Obviously, the comparison with conventional methods is not fair, since G-NLM relies on precious auxiliary data to improve performance.
On the other hand, our first aim is exactly to show that optical-guided despeckling is a simple and safe solution towards obtaining high-quality despeckled SAR data.

\begin{figure}
	\centering\footnotesize
	\begin{tabular}{c@{\hspace{1mm}}c@{\hspace{1mm}}c}
        enhanced-Lee & PPB & SAR-BM3D \\
		\includegraphics[width=2.8cm]{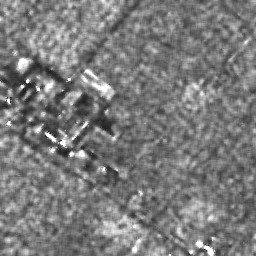} &
		\includegraphics[width=2.8cm]{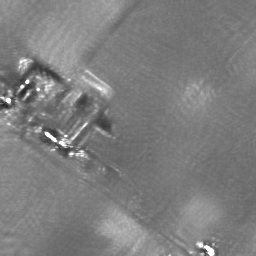} &
		\includegraphics[width=2.8cm]{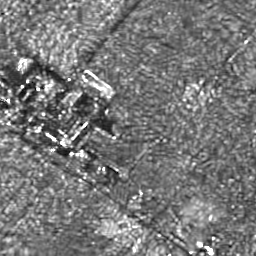} \vspace{1mm} \\
        optical guide & single-look SAR data & FANS \\
		\includegraphics[width=2.8cm]{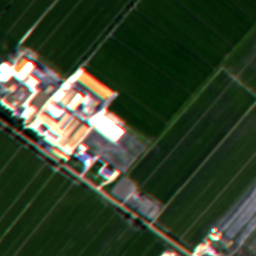} &
		\includegraphics[width=2.8cm]{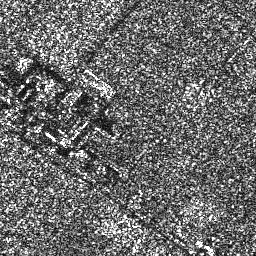} &
		\includegraphics[width=2.8cm]{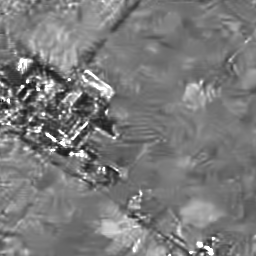} \vspace{1mm} \\
        G-BF & G-NLM sharp & G-NLM smooth \\
		\includegraphics[width=2.8cm]{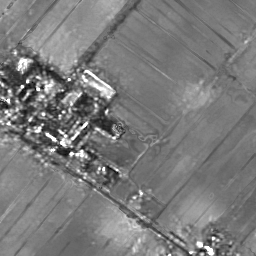} &
		\includegraphics[width=2.8cm]{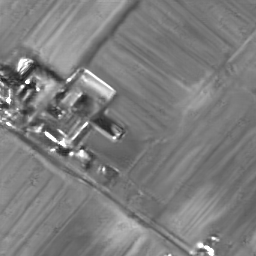} &
		\includegraphics[width=2.8cm]{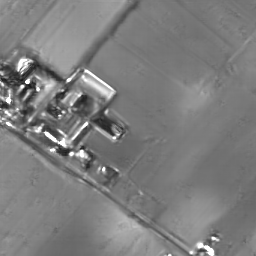} \vspace{1mm} \\
    \end{tabular}
    \caption{Filtering results for the C4 clip. The single-look original is shown in the center for easy comparison.
    G-NLM accurately preserves man-made structures, removing speckle also in their proximity.}
    \label{fig:comparison_C4}
\end{figure}

However, to fully support this claim, we must consider more challenging scenes, man-made structures, roads, and sharp boundaries between regions of different nature.
To this end, in Fig.\ref{fig:comparison_C4} we show a 256$\times$256 detail of the C4 clip.
The fields show the phenomena already described before,
just note that both enhanced Lee and PPB suggest the presence of diagonal strips, further enhanced in G-NLM sharp.
As for the buildings, enhanced Lee and PPB smear and sometimes lose details;
SAR-BM3D and related FANS and G-BF preserve all structures very well, with an accuracy comparable to that of G-NLM.
The latter, however, succeed in removing speckle even inside the man-made area on in its near proximity,
providing a sharper result and contributing to a better perceived quality.

\begin{table}
{\footnotesize
\caption{Structuredness index (RIS) on T-SAR clips.}
\centering
\setlength{\tabcolsep}{4pt}
\begin{tabular}{c|rrrrrrr}
\ru        Clip &   \local  &    \ppb   &    \bmtd  &    \fans  &      \gbf &   \gnlmA  &   \gnlmC  \\ \hline
\ru          T1 &    4.86~  &    5.53~  &    1.45~  &    3.21~  &    3.30~  &    5.84~  &    5.27~  \\
\ru          T2 &    4.42~  &    6.59~  &    2.16~  &    4.94~  &    5.32~  &    6.59~  &    6.61~  \\
\ru          T3 &    5.05~  &    5.80~  &    1.35~  &    2.99~  &    3.54~  &    6.29~  &    5.81~  \\
\ru          T4 &    4.15~  &    6.43~  &    1.94~  &    4.64~  &    3.28~  &    6.62~  &    6.47~  \\ \hline
\ru \i{average} & \i{4.62}~ & \i{6.09}~ & \i{1.72}~ & \i{3.95}~ & \i{3.86}~ & \i{6.33}~ & \i{6.04}~ \\
\end{tabular}                                                                                                                             %
\label{tab:RIS_T}
}
\end{table}

\begin{table}
{\footnotesize
\caption{Structuredness index (RIS) on COSMO clips.}
\centering
\setlength{\tabcolsep}{4pt}
\begin{tabular}{c|rrrrrrr}
\ru        Clip &    \local  &    \ppb   &    \bmtd  &    \fans  &      \gbf &   \gnlmA  &   \gnlmC  \\ \hline
\ru          C1 &     4.63~  &    6.78~  &    2.14~  &    5.19~  &    5.96~  &    6.77~  &    6.83~  \\
\ru          C2 &     4.49~  &    6.69~  &    2.12~  &    5.07~  &    5.41~  &    6.64~  &    6.67~  \\
\ru          C3 &     4.66~  &    6.87~  &    1.87~  &    4.62~  &    5.53~  &    6.90~  &    6.95~  \\
\ru          C4 &     4.68~  &    6.87~  &    1.94~  &    4.90~  &    5.53~  &    6.73~  &    6.69~  \\ \hline
\ru \i{average} &  \i{4.61}~ & \i{6.80}~ & \i{2.02}~ & \i{4.94}~ & \i{5.61}~ & \i{6.76}~ & \i{6.78}~ \\
\end{tabular}
\label{tab:RIS_C}
}
\end{table}

All this said, if we consider our numerical measure of structuredness, the RIS, we obtain quite different indications.
In fact, the results reported in Tab.\ref{tab:RIS_T} and Tab.\ref{tab:RIS_C} for the two datasets
show SAR-BM3D to have by far the lowest RIS, around 2\%, followed by FANS and G-BF (which inherit some good features of SAR-BM3D), and by enhanced Lee,
while both versions of G-NLM come last, together with PPB, with almost 7\%.

\begin{figure}
	\centering\footnotesize
	\begin{tabular}{c@{\hspace{1mm}}c@{\hspace{1mm}}c}
        enhanced-Lee & PPB & SAR-BM3D \\
		\includegraphics[width=2.8cm]{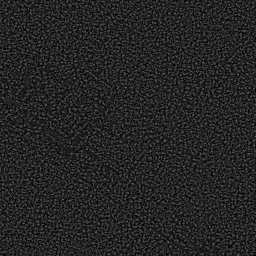} &
		\includegraphics[width=2.8cm]{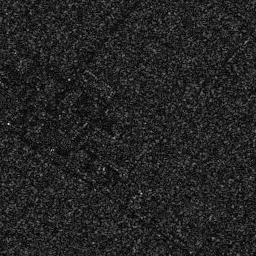} &
		\includegraphics[width=2.8cm]{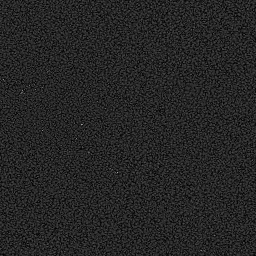} \vspace{1mm} \\
        optical guide & single-look SAR data & FANS \\
		\includegraphics[width=2.8cm]{./Figures/Comparisons/C9_clip_optical.png} &
		\includegraphics[width=2.8cm]{./Figures/Comparisons/C9_clip_SAR.png} &
		\includegraphics[width=2.8cm]{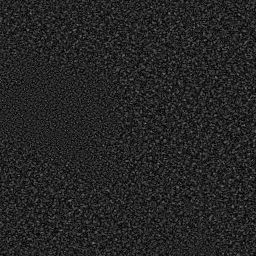} \vspace{1mm} \\
        G-BF & G-NLM sharp & G-NLM smooth \\
		\includegraphics[width=2.8cm]{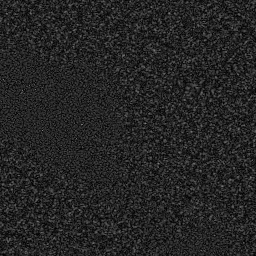} &
		\includegraphics[width=2.8cm]{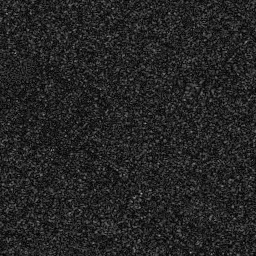} &
		\includegraphics[width=2.8cm]{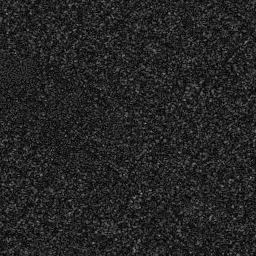} \vspace{1mm} \\
    \end{tabular}
    \caption{Ratio images for the C4 clip. The single-look original is shown in the center for easy comparison.
    Quite limited traces of signal structures are observed (mostly in PPB and G-NLM) while there is an increase in correlation (except for SAR-BM3D).}
    \label{fig:comparison_C4_ratio}
\end{figure}

To explain such conflicting results, in Fig.\ref{fig:comparison_C4_ratio} we show the ratio images themselves.
The visual inspection provides clear answers.
Basically, none of the ratio images show significant leakages from the original image,
some image structures are visible only in the PPB ratio and just barely in the enhanced Lee and G-NLM ratios.
In these conditions,
the RIS measures mostly the grain of the ratio image, which overwhelms truly structural dependencies.
Under this point of view, the SAR-BM3D ratio image is clearly preferable, as it resembles very closely a white noise field.
All other methods introduce some weak correlation which, however, does not seem to impact on image quality.
In summary, RIS provides some valuable information (the ratio image grain) but correlates poorly with image quality.
Unfortunately, this holds for all other measures we tested,
which leaves visual inspection as the most reliable form of quality assessment.

\begin{figure}
	\centering\footnotesize
	\begin{tabular}{c@{\hspace{1mm}}c@{\hspace{1mm}}c}
        enhanced-Lee & PPB & SAR-BM3D \\
		\includegraphics[width=2.8cm]{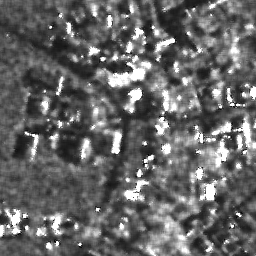} &
		\includegraphics[width=2.8cm]{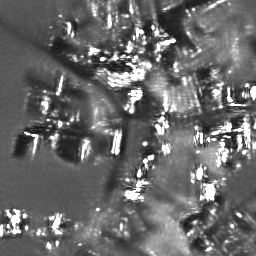} &
		\includegraphics[width=2.8cm]{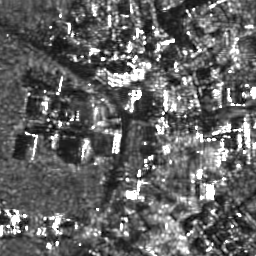} \vspace{1mm} \\
        optical guide & single-look SAR data & FANS \\
		\includegraphics[width=2.8cm]{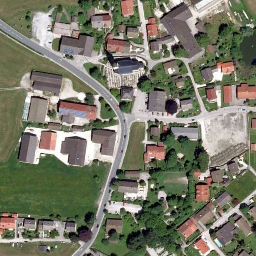} &
		\includegraphics[width=2.8cm]{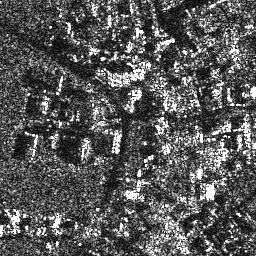} &
		\includegraphics[width=2.8cm]{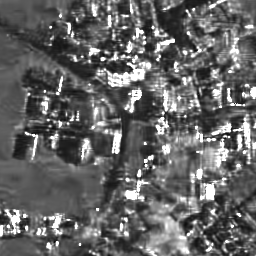} \vspace{1mm} \\
        G-BF & G-NLM sharp & G-NLM smooth \\
		\includegraphics[width=2.8cm]{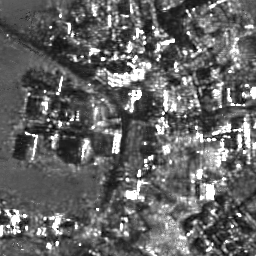} &
		\includegraphics[width=2.8cm]{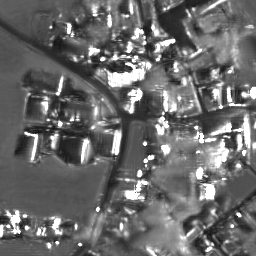} &
		\includegraphics[width=2.8cm]{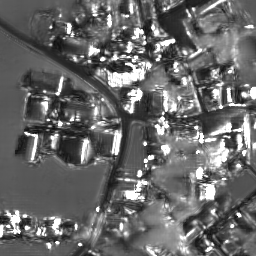} \vspace{1mm} \\
    \end{tabular}
    \caption{Filtering results for the T1 clip. The single-look original is shown in the center for easy comparison.}
    \label{fig:comparison_T1}
\end{figure}

Therefore, we conclude our analysis by studying, with Fig.\ref{fig:comparison_T1}, a last detail, the central part of clip T1, comprising mostly urban areas.
Again, G-NLM preserves faithfully all features related to man-made objects and guarantees a strong rejection of speckle in homogeneous areas, even amidst buildings.
All reference methods, instead, present some shortcomings, like limited speckle rejection, loss of resolution, or the introduction of filtering artifacts.

\begin{figure}[t]
	\centering\footnotesize
	\begin{tabular}{c@{\hspace{1mm}}c@{\hspace{1mm}}c}
        2002 optical guide & 2008 single-look SAR & 2009 optical guide \\
		\includegraphics[width=2.8cm]{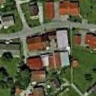} &
		 \includegraphics[width=2.8cm]{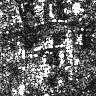} &
		\includegraphics[width=2.8cm]{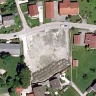} \vspace{1mm} \\
        2002-G-NLM sharp & SAR-BM3D & 2009-G-NLM sharp \\
		\includegraphics[width=2.8cm]{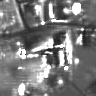} &
		\includegraphics[width=2.8cm]{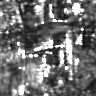} &
		\includegraphics[width=2.8cm]{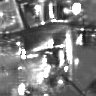} \vspace{1mm} \\
    \end{tabular}
    \caption{Robustness to SAR-optical mismatches.
    SAR-optical inconsistencies (2009 guide) do not disrupt filtering results,
    which remain very similar to those obtained with a better reference (2002 guide) or with conventional filters.}
    \label{fig:comparison_T1_zoom}
\end{figure}

So, the analysis of this image does not seem to add new information.
However, by looking at the whole set of Google Earth images of this area, we discovered some significant temporal changes,
which allow us to study the robustness of the proposed method to mismatches between SAR and optical data.
In particular, as shown in Fig.\ref{fig:comparison_T1_zoom},
a group a buildings was leveled between 30-10-2002, date of the previous available optical image, and 31-12-2009, date of our guide.
In the test SAR image, dated 27-01-2008, the buildings were still standing, as testified by several double reflection lines.
Therefore, there is strong mismatch between SAR data and optical guide.
Nonetheless, this does not seem to affect the filtered image,
where the building-related structures are clearly visible, and look very similar to those provided by other filters, e.g., SAR-BM3D.
As a further proof, we applied the proposed filter using the 2002 image as optical guide, and obtaining similar results.

\section{Conclusions}
In this paper, we proposed a nonlocal SAR despeckling filter which makes use of available optical imagery to improve performance.
Experiments on two real-world datasets show the proposed method to provide filtered images of excellent quality,
arguably out of the reach of purely SAR-domain methods.
The performance is also much better than that of our own previous optical-guided filter.

It is not surprising that information provided by optical imagery may help improving SAR despeckling.
Patch-wise nonlocal filtering allows us to exploit this information in a seamless way, avoiding any optical-induced artifacts.
However, better solutions are certainly possible, and we hope to witness increasing activity on this line of research.
A crucial point along this path is user-friendliness.
End users willing to obtain high-quality despeckling images look for simple and efficient plug-and-play tools,
which do not require much direct involvement.

The proposed method moves a step in this direction.
It provides high-quality and stable results based on the freely available Google Earth images, even in the presence of temporal changes.
However, the user is still required to manually co-register the optical images with their SAR data.
Our future work aims at improving the automation of this latter phase.

\newpage
\section*{\noindent Appendix A}

In this appendix we compute distribution, mean and variance of the random variable
\begin{equation}
    D = \log \left[ \frac{X+Y}{2\sqrt{XY}} \right] 
\end{equation}
where $X$ and $Y$ are i.i.d. RV's, with unit-mean Gamma distribution,
\begin{equation}
    p_X(x) = \frac{L^L}{\gammaf{L}} x^{L-1}e^{-Lx}1(x)
\end{equation}
which model speckle samples in a $L$-look SAR image.

Since $X$ and $Y$ are non negative, their geometric and arithmetic means, $G=\sqrt{XY}$ and $A=(X+Y)/2$, are well defined and non negative,
and their ratio does not exceed 1, $R=G/A \in [0,1]$.
Then, we consider the RV transformation
\begin{equation}
    \left\{ \begin{array}{l}
        A = (X+Y)/2 \\
        R = 2\sqrt{XY}/(X+Y) \rule{0mm}{5mm}
            \end{array} \right.
\end{equation}
with inverse transformation
\begin{equation}
    \left\{ \begin{array}{l}
        X = A(1\pm\sqrt{1-R^2}) \\
        Y = A(1\mp\sqrt{1-R^2}) \rule{0mm}{5mm}
            \end{array} \right.
\end{equation}
and Jacobian ${\partial(x,y)}/{\partial(a,r)}$ with determinant
\begin{equation}
    \left| \begin{array}{cc}
        \partial x/\partial a & \partial x/\partial r \\
        \partial y/\partial a & \partial y/\partial r \end{array} \right| = \frac {2ar}{\sqrt{1-r^2}}
\end{equation}
by which we obtain
\begin{IEEEeqnarray}{rCl}
    p_{AR}(a,r) & = & p_X(a(1\pm\sqrt{1 \minus r^2})) \times \nonumber \\
                &   & p_Y(a(1\mp\sqrt{1 \minus r^2})) \left| \frac{\partial(x,y)}{\partial(a,r)} \right| \\
                & = & 2 \frac{L^{2L}}{\gammaf{L}^2} (a^2r^2)^{L-1} e^{-2La} \frac{2ar}{\sqrt{1 \minus r^2}} 1(a)\Pi(r) \nonumber
\end{IEEEeqnarray}
having defined $\Pi(r)=1(r)-1(r-1)$.

By rearranging the terms of the above expression we see that $A$ and $R$ are independent random variables
\be
    p_{AR}(a,r) = p_A(a) p_R(r)
\ee
where $A$ is a unit-mean Gamma RV with parameter $2L$
\be
    p_A(a) = \frac{(2L)^{2L}}{\gammaf{2L}} a^{2L-1}e^{-2La}1(a)
\ee
while the ratio $R$ has pdf
\be
    p_R(r) = \frac{\gammaf{2L}}{[2^{L-1}\gammaf{L}]^2} \frac{r^{2L-1}}{\sqrt{1-r^2}} \Pi(r)
\ee
A further RV transformation $D=-\log(R)$ provides the desired pdf
\be
    p_D(d) = C(L) \frac{e^{-2Ld}}{\sqrt{1-e^{-2d}}}1(d)
\ee
with $C(L)=\gammaf{2L}/[2^{L-1}\gammaf{L}]^2$.

To obtain mean and variance of $D$ we compute its moment generating function
\begin{IEEEeqnarray}{rCl}
    M_D(s) & = & E[e^{sD}] = \int_0^\infty C(L) \frac{e^{-2(L \minus s/2)d}}{\sqrt{1-e^{-2d}}}dd \\
           & = & \frac{C(L)}{C(L \minus s/2)} =
                 \frac{\gammaf{2L}}{\gammaf{2L \minus s}} \left[\frac{\gammaf{L \minus s/2}}{\gammaf{L}}\right]^2 2^{-s} \nonumber
\end{IEEEeqnarray}
with derivatives
\be
    M'_D(s) = M_D(s)\left[ \psif{0}{2L \minus s} - \psif{0}{L \minus \frac{s}{2}} - \log2 \right]
\ee
and
\begin{IEEEeqnarray}{rCl}
    M''_D(s) & = & [M'_D(s)]^2/M_D(s) + \\
             &   & + M_D(s)\left[ \frac{1}{2}\psif{1}{L \minus s/2} - \psif{1}{2L \minus s} \right] \nonumber
\end{IEEEeqnarray}
expressed in terms of the $m$-order polygamma functions
\be
    \psif{m}{x} = \frac{d^{m+1}\log[\gammaf{x}]}{dx^{m+1}}
\ee
Therefore
\be
    E[D] = M'_D(0) = \psif{0}{2L} - \psif{0}{L} - \log2
\ee
and
\begin{IEEEeqnarray}{rCl}
    {\rm VAR}[D] & = & M''_D(0) - [M'_D(0)]^2 \\
                 & = & \frac{1}{2}\psif{1}{L} - \psif{1}{2L} \nonumber
\end{IEEEeqnarray}

\balance
\bibliographystyle{IEEEtran}
{\footnotesize \bibliography{references}}

\end{document}